\documentclass[letterpaper]{article} 
\usepackage{aaai24}  
\usepackage{times}  
\usepackage{helvet}  
\usepackage{courier}  
\usepackage[hyphens]{url}  
\usepackage{graphicx} 
\usepackage{natbib}  
\usepackage{caption} 
\usepackage{algorithm}
\usepackage{algorithmic}
\usepackage{amsmath}
\usepackage{amsfonts}
\usepackage{bm}
\usepackage{multirow}
\usepackage{multicol}
\usepackage{booktabs}
\usepackage[table]{xcolor}
\usepackage{subcaption}

\def\eg{\emph{e.g., }}

\usepackage{newfloat}
\usepackage{listings}
\DeclareCaptionStyle{ruled}{labelfont=normalfont,labelsep=colon,strut=off} 
\floatstyle{ruled}
\newfloat{listing}{tb}{lst}{}
\floatname{listing}{Listing}
\title{VSFormer: Visual-Spatial Fusion Transformer for Correspondence Pruning}
\author{
    Tangfei Liao\textsuperscript{\rm 1}, 
    Xiaoqin Zhang\textsuperscript{\rm 1}\thanks{Corresponding authors}, 
    Li Zhao\textsuperscript{\rm 1}, 
    Tao Wang\textsuperscript{\rm 2}, 
    Guobao Xiao\textsuperscript{\rm 3}\footnotemark[1]
}
\affiliations{
    \textsuperscript{\rm 1}Key Laboratory of Intelligent Informatics for Safety and Emergency of Zhejiang Province, Wenzhou University, China \\
    \textsuperscript{\rm 2}State Key Lab for Novel Software Technology, Nanjing University, China \\
    \textsuperscript{\rm 3}School of Electronics and Information Engineering, Tongji University, China \\

    \{tangfeiliao, zhangxiaoqinnan, taowangzj\}@gmail.com, 
    lizhao@wzu.edu.cn, x-gb@163.com
}

\usepackage{bibentry}

\begin{document}

\maketitle

\begin{abstract}
Correspondence pruning aims to find correct matches (inliers) from an initial set of putative correspondences, which is a fundamental task for many applications. 
The process of finding is challenging, given the varying inlier ratios between scenes/image pairs due to significant visual differences. 
However, the performance of the existing methods is usually limited by the problem of lacking visual cues (\eg texture, illumination, structure) of scenes. 
In this paper, we propose a Visual-Spatial Fusion Transformer (VSFormer) to identify inliers and recover camera poses accurately. 
Firstly, we obtain highly abstract visual cues of a scene with the cross attention between local features of two-view images. 
Then, we model these visual cues and correspondences by a joint visual-spatial fusion module, simultaneously embedding visual cues into correspondences for pruning. 
Additionally, to mine the consistency of correspondences, we also design a novel module that combines the KNN-based graph and the transformer, effectively capturing both local and global contexts. 
Extensive experiments have demonstrated that the proposed VSFormer outperforms state-of-the-art methods on outdoor and indoor benchmarks. 
Our code is provided at the following repository: https://github.com/sugar-fly/VSFormer. 
\end{abstract}

\section{Introduction}

\begin{figure}[t]
\begin{center}
    \includegraphics[width=0.9\linewidth]{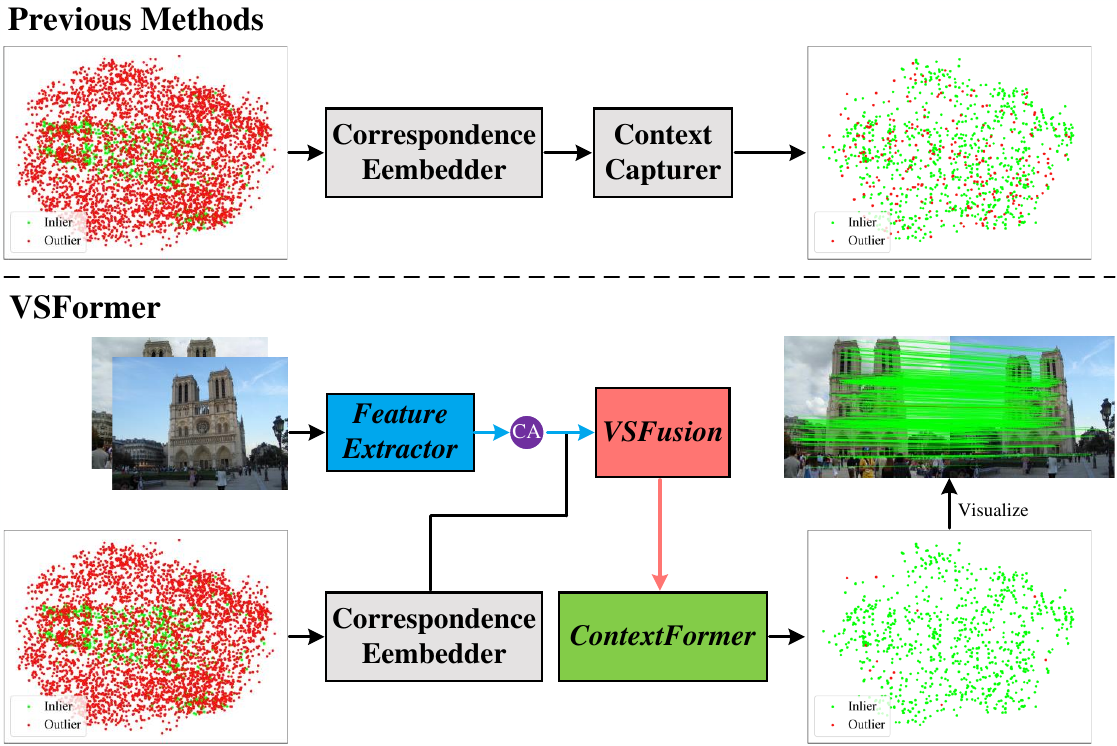}
    \caption{Comparison between previous methods and ours. Top: the architecture of previous methods, which lack the visual perception of a scene. 
    Bottom: the architecture of our VSFormer introduces visual cues of a scene to guide correspondence pruning. 
    For visualization purposes, the correspondences (4D) across two-view images are projected into a 2D space by t-SNE \cite{Van2008}. 
    The circle CA represents the cross-attention layer. 
    }\label{figure1}
\end{center}
\end{figure} 

Two-view correspondence learning aims to establish reliable correspondences/matches across two images and accurately recover camera poses, which is a fundamental task in computer vision and plays an important role in many applications such as simultaneous localization and mapping~\cite{mur2015orb}, structure from motion~\cite{schonberger2016structure} and image registration~\cite{xiao2020deterministic}. 
However, the outlier (false match) ratio in initial correspondences is often over 90\% due to various cross-image variations (\eg low texture, illumination changes, repetitive structures), which severely undermines the performance of downstream tasks. 
Therefore, much recent research has focused on pruning false matches from initial correspondences to obtain accurate two-view geometry. 

Some traditional methods such as RANSAC~\cite{Fischler1981} and its variants~\cite{Chum2005a, Barath2019} search for correct correspondences (inliers) using iterative sampling strategies, but their running time grows exponentially with outlier ratio, thus making them unsuitable for tackling high-outlier problems. 
Meanwhile, learning-based methods have achieved promising performance. 
PointCN~\cite{Yi2018} is a pioneering work, handling the disordered property of correspondences (visualization in Fig.~\ref{figure1}) with the multilayer perceptron (MLP) architecture. 

Until recently, the most popular networks commonly employed an iterative network to inherit weights from the previous iteration, which greatly improved the performance of correspondence pruning. 
These iterative networks typically designed some extra structures to mine the geometric consistency within correspondences, such as OANet~\cite{Zhang2019}, ACNet~\cite{Sun2020}, MS$^2$DG-Net~\cite{Dai2022}. 
While such a direction deserves further exploration, these methods only considered spatial information of correspondences as input, which significantly hindered the acquisition of deep information and simultaneously damaged the network performance. 
Thus, there are some researches~\cite{Luo2019, Liu2021} that employ feature point descriptors of a single image to enhance the representation ability of network inputs. 
In this paper, we lean toward another perspective and ask the following question: 
\textit{Can we provide the network with a scene-aware prior at a higher level to guide pruning?} 
That is, if the inlier ratio of a scene/image pair can be perceived in advance, it will facilitate the network in discriminating some ambiguous correspondences. 

To this end, with the observation that the inlier ratio varies greatly between scenes/image pairs due to significant visual differences (\eg texture, illumination, occlusion), we adopt some scene visual cues as an abstract representation of the inlier ratio. 
As shown in Fig.~\ref{figure1}, compared to previous methods, we add some steps for extracting and fusing scene visual cues. 
Specifically, the proposed Visual-Spatial Fusion Transformer (VSFormer) is mainly composed of three components: Visual Cues Extractor (VCExtractor), Visual-Spatial Fusion (VSFusion) module, and Context Transformer (ContextFormer). 
Firstly, the VCExtractor extracts scene visual cues with the cross attention between local features of two-view images. 
Then, a novel Visual-Spatial Fusion (VSFusion) module is designed to model the relationship between visual cues and spatial cues, simultaneously embedding visual cues into correspondences. 
The VSFusion involves two phases: 
\romannumeral1) the module adopts a transformer~\cite{Vaswani2017} to model the complex intra- and inter-modality relationships of visual and spatial cues; 
\romannumeral2) the module encodes visual and spatial cues separately, using a simple element-wise summation operation to fuse them; 
Meanwhile, to facilitate fusion, VSFusion uses soft assignment manner~\cite{Zhang2019} to project spatial cues into the same space as visual cues. 
The proposed VSFusion effectively embeds scene visual cues into correspondences for guiding subsequent correspondence pruning. 

To fully mine contextual information of correspondences, we also propose a novel structure called ContextFormer for pruning, simply stacking a transformer sub-network on top of a graph neural network (GNN). 
In GNN, a novel graph attention block is designed to improve the representation ability of a KNN-based graph. 
The block adopts the squeeze-and-excitation mechanism to efficiently capture the potential spatial-, channel-, and neighborhood-wise relations inside a KNN-based graph, facilitating neighborhood aggregation. 
The proposed structure exploits the neighborhood information of a KNN-based graph and the global modeling ability of the transformer, explicitly capturing both local and global contexts of correspondences, thus further improving the performance of our method. 

The contributions of this paper are summarized as follows: 
\begin{itemize}
	\item A visual-spatial fusion transformer is proposed to extract and embed scene visual cues into correspondences for guiding pruning. Meanwhile, we design a joint visual-spatial fusion module to fuse visual and spatial cues in the same space. To the best of our knowledge, this is the first time that scene cues have been introduced for correspondence pruning. 
	\item A simple yet effective ContextFormer is proposed to explicitly capture both local and global contexts of correspondences. In this structure, we also design a graph attention block based on the squeeze-and-excitation mechanism to enhance the representation ability of a KNN-based graph. 
	\item The proposed VSFormer achieves a precision increase of 15.79\% and 4.45\% compared with the state-of-the-art result on outdoor and indoor benchmarks, respectively. 
\end{itemize}

\section{Related Work}
\label{sec:related}
\subsubsection{Traditional Methods.}
RANSAC~\cite{Fischler1981} and its variants based on iterative sampling strategies to estimate a geometry model. 
To be specific, these methods resample the smallest subset of input correspondences to estimate a parametric model as a hypothesis, and then verify its confidence by counting the number of consistent inliers. 
PROSAC~\cite{Chum2005a} can significantly expedite this process. 
USAC~\cite{Raguram2012} proposes a unified framework that incorporates multiple advancements for RANSAC variants. 
MAGSAC~\cite{Barath2019} uses $\sigma$-consensus to eliminate the requirement for a predefined inlier-outlier threshold. 
RANSAC and its variants have been widely recognized as the standard solution for robust model estimation. 
However, their performance degrades severely as the outlier ratio increases~\cite{ma2021image}. 

\subsubsection{Learning-Based Methods.}
PointCN~\cite{Yi2018} is a pioneering work that formulates correspondence pruning as both an essential matrix regression problem and a binary classification problem. 
It employs MLPs to effectively process the disordered property of correspondences and proposes a context normalization technique to embed global information into each correspondence. 
DFE~\cite{Ranftl2018} proposes a distinct loss function and an iterative network based on PointCN. 
ACNe~\cite{Sun2020} employs the attention mechanism to enhance the performance of the network. 
OANet~\cite{Zhang2019} performs full-size prediction for all initial correspondences and introduces a clustering layer to capture local context. 
MSA-Net~\cite{Zheng2022} and PGFNet~\cite{Liu2023} also propose some jointly spatial-channel attention blocks to capture the global context of correspondences. 
After that, there are some researches based on the graph neural network~\cite{Zhao2021, Dai2022, liao2023sga}. 
CLNet~\cite{Zhao2021} introduces a neighborhood aggregation manner and the pruning strategy to refine coarse correspondences. 
MS$^2$DG-Net~\cite{Zhao2021} builds KNN-based graphs at different stages and employs the multi-head self-attention mechanism to enhance the representation ability of graphs. 
Although these methods have achieved promising performance, they only consider spatial information of correspondences as input, which severely limits the acquisition of deep information and simultaneously impairs network performance. 
Therefore, LMCNet~\cite{Liu2021} exploits feature point descriptors of a single image to enhance the representation ability of correspondences. 
In this paper, with another perspective, jointly visual-spatial cues as a scene-aware prior to guide correspondence pruning. 

\section{Method}
\label{sec:method}
\subsection{Problem Formulation}
Given two-view images $(\mathbf{I_A}, \mathbf{I_B})$, our task is to precisely identify correct correspondences from initial correspondences and recover camera poses. 
To be specific, feature points and descriptors are first extracted from two-view images using a feature detector (\eg SIFT~\cite{Lowe2004} and SuperPoint~\cite{DeTone2018}). 
Then, the initial correspondence set $\mathbf{I_C}$ is built by a nearest neighbor matching strategy: 
\begin{equation} \label{eq1}
    \mathbf{I_C} = \mathbf{\left\{ c_1,c_2,...,c_N \right\}} \in \mathbb{R}^{N \times 4},~\mathbf{c_i} = (x_i, y_i, x_{i}^{\prime}, y_{i}^{\prime}),
\end{equation}
where $\mathbf{c_i}$ is the $i$-th correspondence; $(x_i, y_i)$ and $(x_{i}^{\prime}, y_{i}^{\prime})$ are the feature point coordinates of the given two-view images that have been normalized by camera intrinsics~\cite{Zhang2019}. 

In our task, the correspondence pruning is typically formulated as an essential matrix regression problem and an inlier/outlier classification problem~\cite{Yi2018}. 
Following CLNet~\cite{Zhao2021}, this paper iteratively uses ContextFormer for correspondence pruning and produces the final probability set $\mathbf{P_f = \left\{ p_1,...,p_i,...,p_{N^{'}} \right\}}$ of candidates, which indicates the probability of each candidate as an inlier. The above process can be formulated as follows: 
\begin{gather} \label{eq2}
    \mathbf{(C_f, W_f)} = f_{\phi}(\mathbf{I_A, I_B, I_C}), \\
    \mathbf{P_f} = \mathrm{Softmax}(\mathbf{W_f}), 
\end{gather}
where $\mathbf{W_f = \left\{ w_1,...,w_i,...,w_{N^{'}} \right\}}$ represents the weights of final candidates; 
$\mathbf{C_f = \left\{ c_1,...,c_i,...,c_{N^{'}} \right\}}$ represents the final candidate set; 
$f_{\phi}(\cdot)$ indicates our proposed VSFormer; 
${\phi}$ indicates the network parameters. 

Then, the final candidate set $\mathbf{C_f}$ and the probability set $\mathbf{P_f}$ are taken as input, and a weighted eight-point algorithm~\cite{Yi2018} is applied to regress the essential matrix. The process is presented as: 
\begin{equation} \label{eq3}
    \mathbf{\widehat{E}} = g(\mathbf{C_f}, \mathbf{P_f}),
\end{equation}
where $g(\cdot, \cdot)$ represents a function of the weighted eight-point algorithm, and the matrix $\mathbf{\widehat{E}}$ indicates the predicted essential matrix. 

In addition, following~\cite{Zhao2021}, this paper also adopts the full-size verification approach to deal with the inlier/outlier classification problem. 
Specifically, the matrix $\mathbf{\widehat{E}}$ and the initial correspondence set $\mathbf{I_C}$ are taken as inputs to produce the predicted symmetric epipolar distance set $\mathbf{\widehat{D}}$. 
Note that an empirical threshold ($10^{-4}$) of epipolar distance is used criterion to discriminate outliers from inliers~\cite{Hartley2003}. 
The process can be formulated as follows: 
\begin{equation} \label{eq4}
    \mathbf{\widehat{D}} = h(\mathbf{\widehat{E}, I_C}),
\end{equation}
where $h(\cdot, \cdot)$ represents a function of the full-size verification. 

\begin{figure}[t]
\begin{center}
    \includegraphics[width=\linewidth]{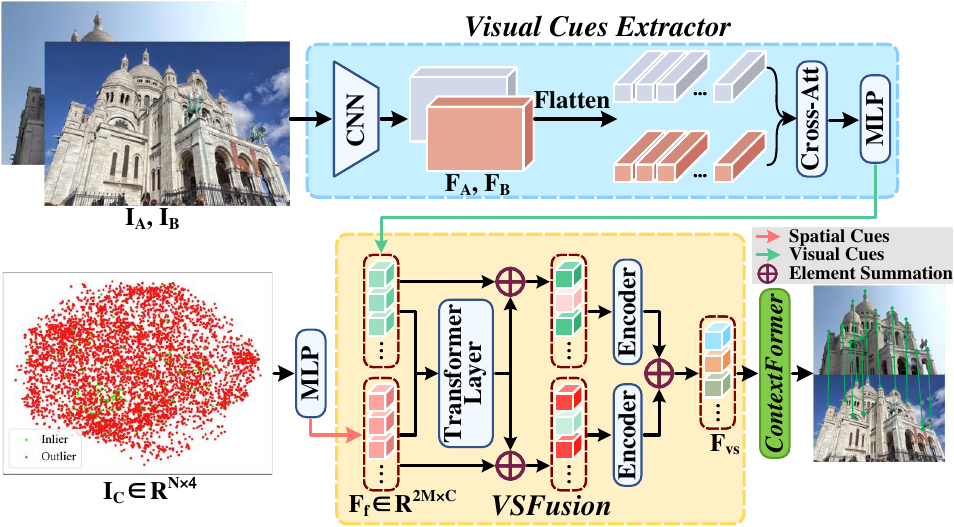}
    \caption{The architecture of our VSFormer mainly contains Visual Cues Extractor (VCExtractor), Visual-Spatial Fusion (VSFusion) Module, and Context Transformer (ContextFormer). Note that we omit the inlier predictor after ContextFormer for simplicity. }\label{figure2}
\end{center}
\end{figure} 

\subsection{Visual-Spatial Fusion Module}
On the one hand, the vast majority of methods only use the spatial information of correspondences as input, which is still challenging for datasets with a large number of outliers. 
On the other hand, some researches~\cite{Luo2019, Liu2021} employ feature point descriptors of a single image to further improve the network performance. 
However, existing methods lack a scene-aware prior to guide correspondence pruning. 
In this paper, we first base on the fact that the inlier ratio varies greatly between scenes/image pairs due to significant visual differences (\eg texture, illumination, occlusion). 
Then, we extract some visual cues of a scene/image pair to abstractly represent the inlier ratio, which is beneficial for the network to distinguish some ambiguous correspondences. 
To this end, as illustrated in Fig.~\ref{figure2}, we propose Visual Cues Extractor (VCExtractor) and Visual-Spatial Fusion (VSFusion) module for extracting and fusing visual cues into correspondences. 

\subsubsection{Visual Cues Extractor.}
VCExtractor is used to extract scene visual cues, and when significant visual differences such as occlusions, large viewpoint changes, and illumination changes between two-view images, the attention map scores in the cross-attention layer tend to be generally low. This message is passed through visual cues to the VSFusion and embedded into each correspondence. 
To be specific, in our VCExtractor, a standard convolution architecture with ResNet34~\cite{He2016} is first used to extract high-dimensional local features $\mathbf{\left\{ F_A, F_B \right\}} \in \mathbb{R}^{C_F \times \frac{H}{4} \times \frac{W}{4}}$ from two-view images $\mathbf{\left\{ I_A, I_B \right\}} \in \mathbb{R}^{3 \times H \times W}$. 
Then, local features are flattened into 1-D vectors and delivered into the cross-attention layer~\cite{Sun2021} to produce initial visual cues of a scene. 
Subsequently, these initial visual cues are embedded with an MLP to obtain visual cues $\mathbf{F_v} \in \mathbb{R}^{M \times C}$ for fusing. 
In addition, the initial correspondences $\mathbf{I_C} \in \mathbb{R}^{N \times 4}$ are also embedded with an MLP to extract the deep feature $\mathbf{F_s} \in \mathbb{R}^{N \times C}$ as spatial cues. 

\subsubsection{Visual-Spatial Fusion.}
The VSFusion is responsible for fusing visual and spatial cues in the same space, and projects jointly visual-spatial cues into the original space. 
Firstly, since the fusion between two modalities is beneficial in the same space, VSFusion projects spatial cues into a unified space with visual cues through a learnable soft assignment manner~\cite{Zhang2019}. 
The above process can be formulated as follows: 
\begin{gather} \label{eq5}
    \mathbf{F_{s}^{'}} = (\mathbf{W})^{T} \mathbf{F_{s}}, 
\end{gather}
where $\mathbf{W} \in \mathbb{R}^{N \times M}$ is a learnable matrix. 

Then, a transformer is adopted to robustly model the relationship between visual cues and spatial cues. 
The transformer layer takes the concatenated feature $\mathbf{F_{f}} \in \mathbb{R}^{2M \times C}$ of visual and spatial cues as input. 
For each head in the multi-headed self-attention layer, three learnable matrices $\mathbf{W_Q}$, $\mathbf{W_K}$ and $\mathbf{W_V}$ project the concatenated feature to query $\mathbf{Q} \in \mathbb{R}^{2M \times d}$, key $\mathbf{K} \in \mathbb{R}^{2M \times d}$ and value $\mathbf{V} \in \mathbb{R}^{2M \times d}$, where $d=C/h$ and $h$ is the number of heads. 
Subsequently, the attention matrix $\mathbf{A} \in \mathbb{R}^{2M \times 2M}$ is calculated by applying a row-wise softmax function on $\mathbf{Q K}^T$. 
The messages $\mathbf{F_{f}^{'}} \in \mathbb{R}^{2M \times C}$ are formulated as $\mathbf{A V}$, which fuse the complex relations between visual and spatial cues. 
After that, these messages are processed through a feed-forward network and split into $\mathbf{F_{v}^{'}} \in \mathbb{R}^{M \times C}$ and $\mathbf{F_{s}^{''}} \in \mathbb{R}^{M \times C}$. 
The above process can be simply described as: 
\begin{gather} \label{eq6}
    \mathbf{F_{f}^{'}} = \mathrm{MHSA}(\mathbf{F_{f}}), \\
    \mathbf{(F_{v}^{'}, F_{s}^{''})} = \mathrm{Split}(\mathrm{FFN}(\mathbf{F_{f}^{'}})), 
\end{gather}
where $\mathrm{MHSA(\cdot)}$ denotes the multi-headed self-attention layer described above; $\mathrm{FFN(\cdot)}$ represents the feed-forward network. 

Next, an element-wise summation is employed to obtain jointly visual-spatial cues $\mathbf{F_{vs}} \in \mathbb{R}^{M \times C}$. 
Meanwhile, skip connections and resnet-like encoders~\cite{Yi2018} are used to rebuild the intra-modality context.
The above process can be expressed as: 
\begin{gather} \label{eq7}
    \mathbf{F_{vs}} = \mathrm{R_1}(\mathbf{F_{v} + F_{v}^{'}}) + \mathrm{R_2}(\mathbf{F_{s}^{'} + F_{s}^{''}}), 
\end{gather}
where $\mathrm{R_1(\cdot)}$ and $\mathrm{R_2(\cdot)}$ represent the encoders with different parameters. 
Finally, similar to Eq.~\ref{eq5}, the jointly visual-spatial cues $\mathbf{F_{vs}} \in \mathbb{R}^{M \times C}$ is projected into the original space $\mathbf{F_{vs}^{'}} \in \mathbb{R}^{N \times C}$. 

\begin{figure}[t]
\begin{center}
    \includegraphics[width=\linewidth]{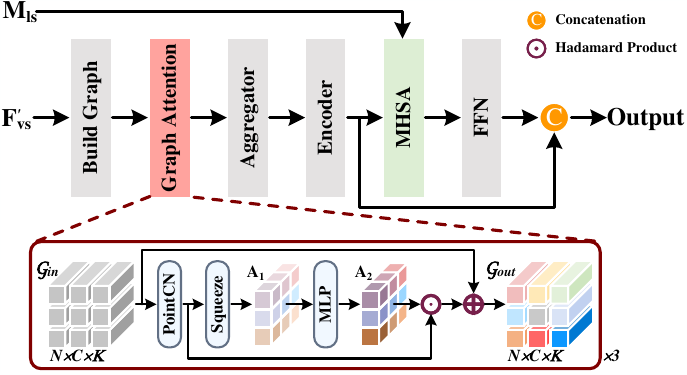}
    \caption{Illustration of our proposed ContextFormer. Meanwhile, we design a novel graph attention block to mine potential relationships along three different dimensions. }\label{figure3}
\end{center}
\end{figure} 

\subsection{Context Transformer}
In our task, mining the consistency within correspondences is important to search for correct matches. 
In this paper, to fully capture contextual information of joint visual-spatial correspondences, a Context Transformer (ContextFormer) structure is designed. 
Intuitively, correct correspondences should be consistent in both their local and global contexts, thus ContextFormer explicitly captures local and global contexts by stacking the graph neural network and transformer, as shown in Fig.~\ref{figure3}. 

\begin{figure*}
    \centering
    \subcaptionbox*{Plain image pair}[.245\linewidth]{
        \includegraphics[width=\linewidth]{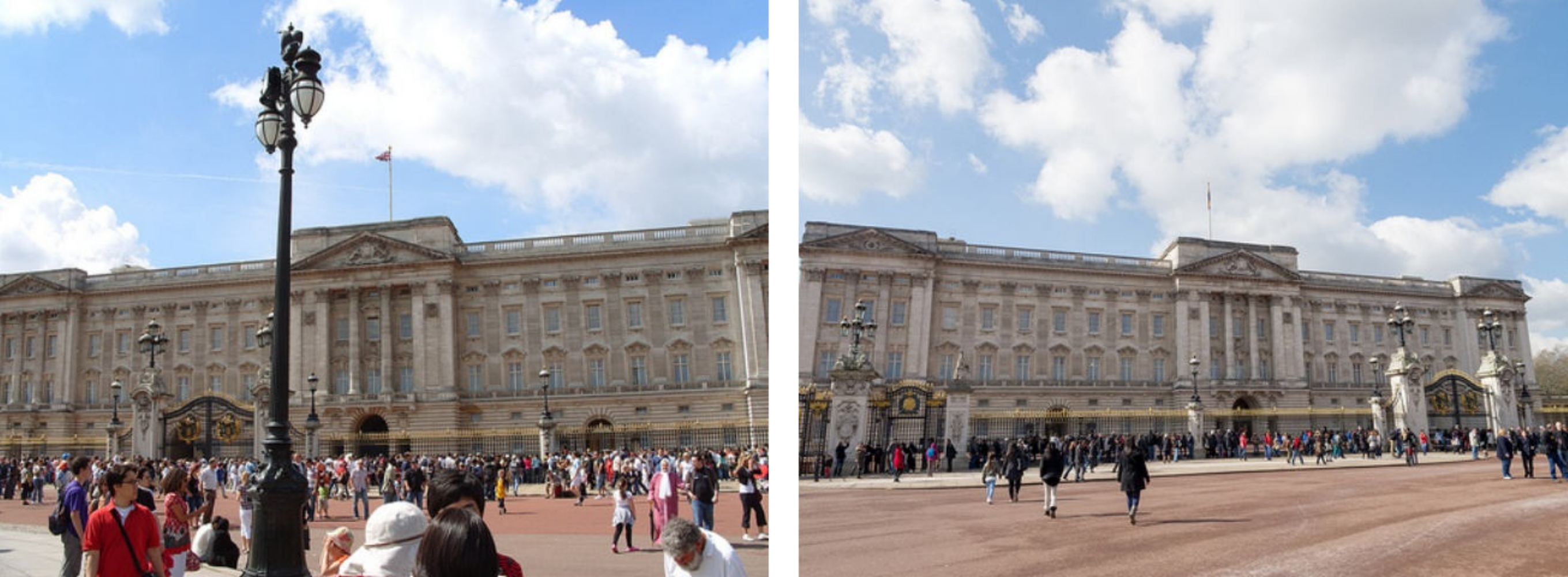}
        \includegraphics[width=\linewidth]{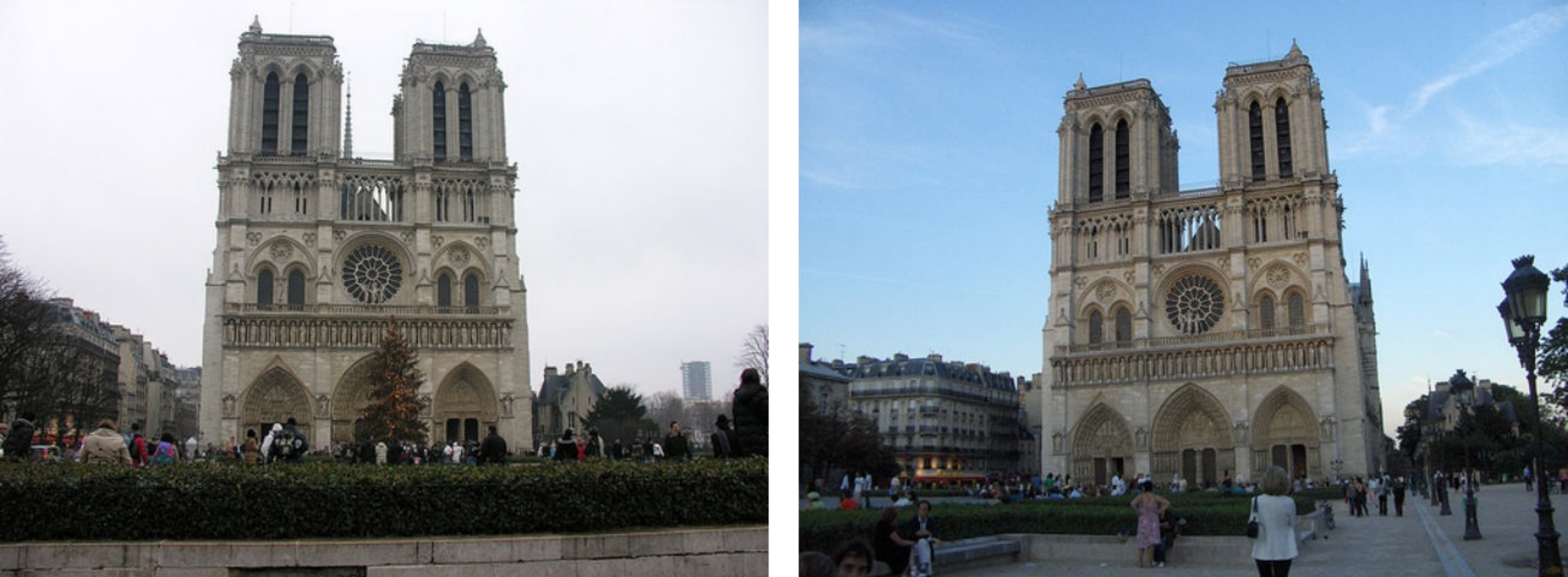}
        \includegraphics[width=\linewidth]{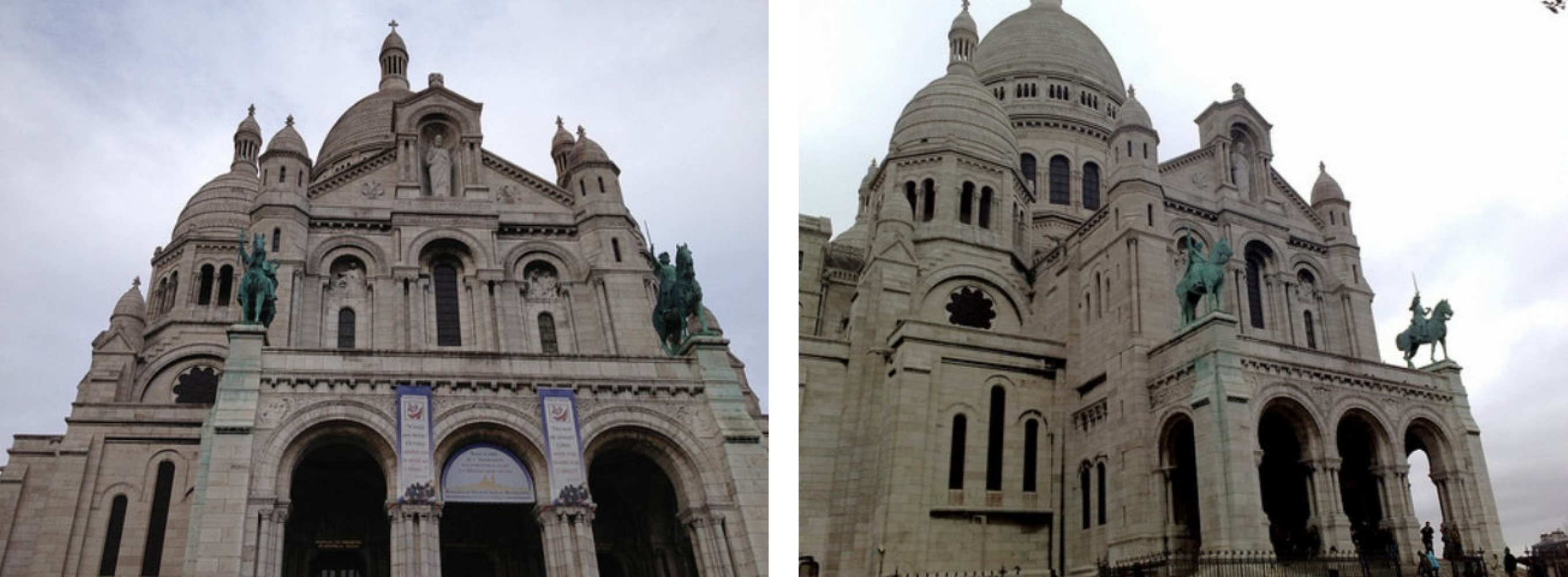}
        \includegraphics[width=\linewidth]{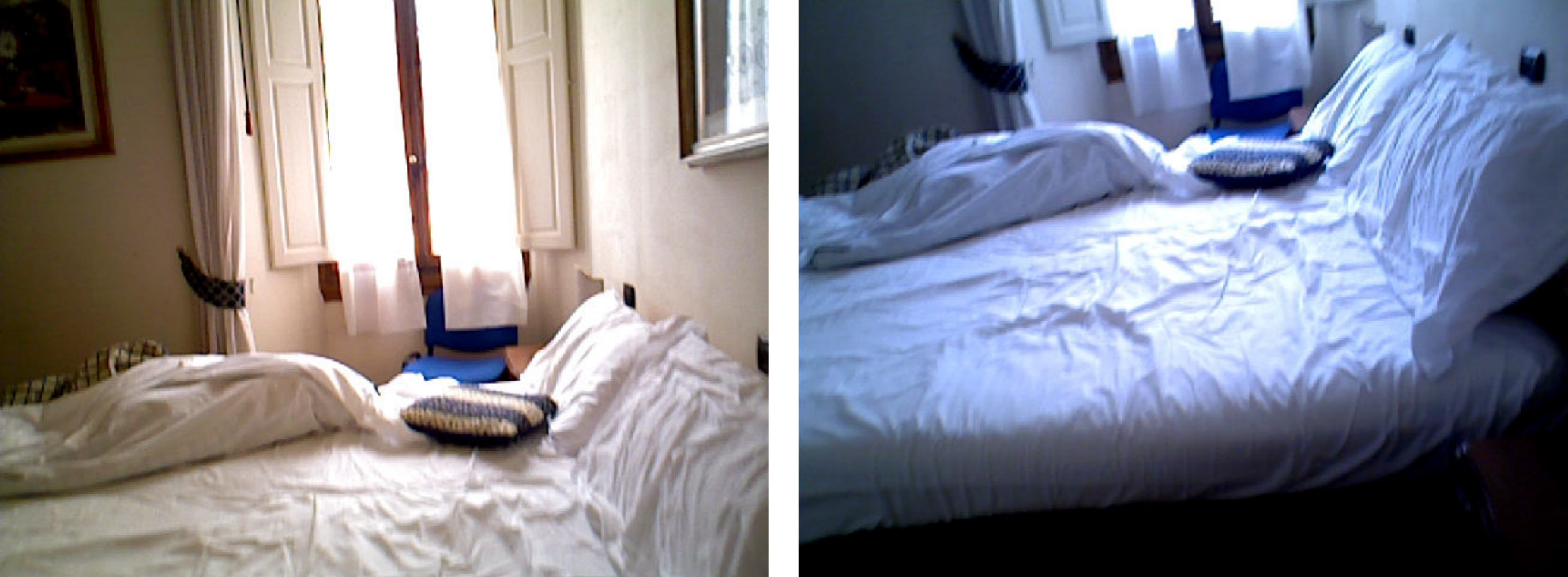}
        \includegraphics[width=\linewidth]{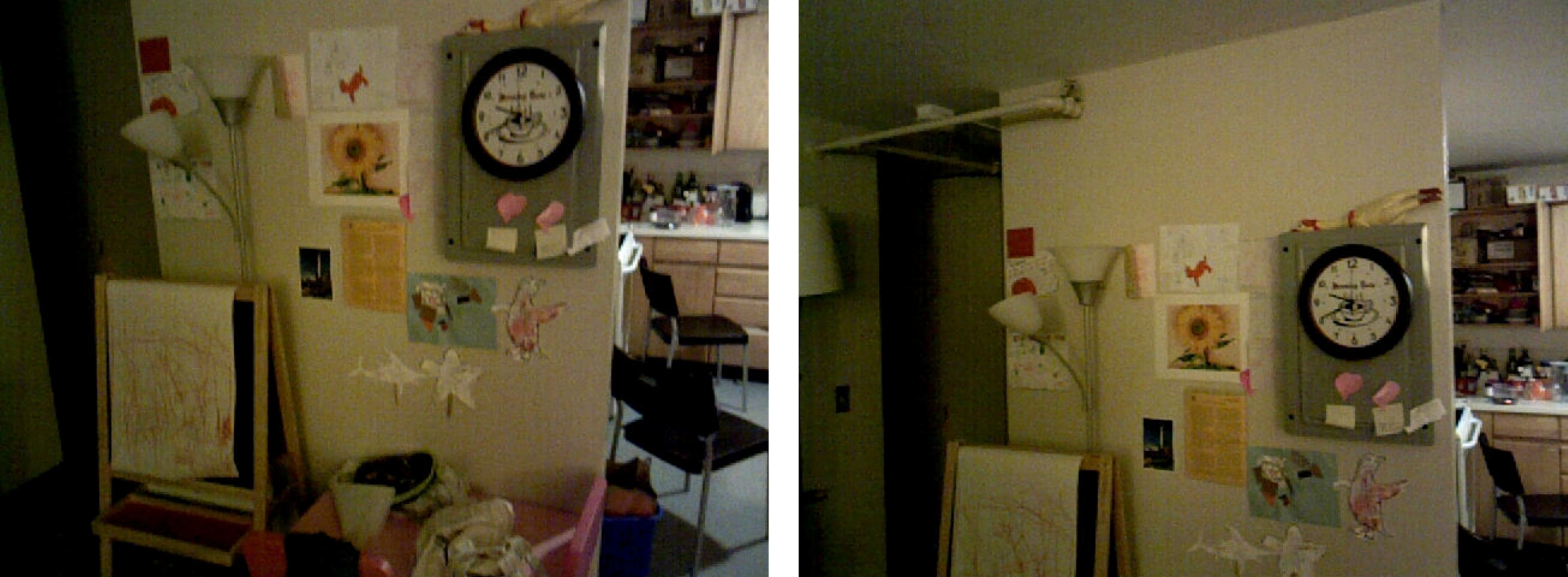}
        \includegraphics[width=\linewidth]{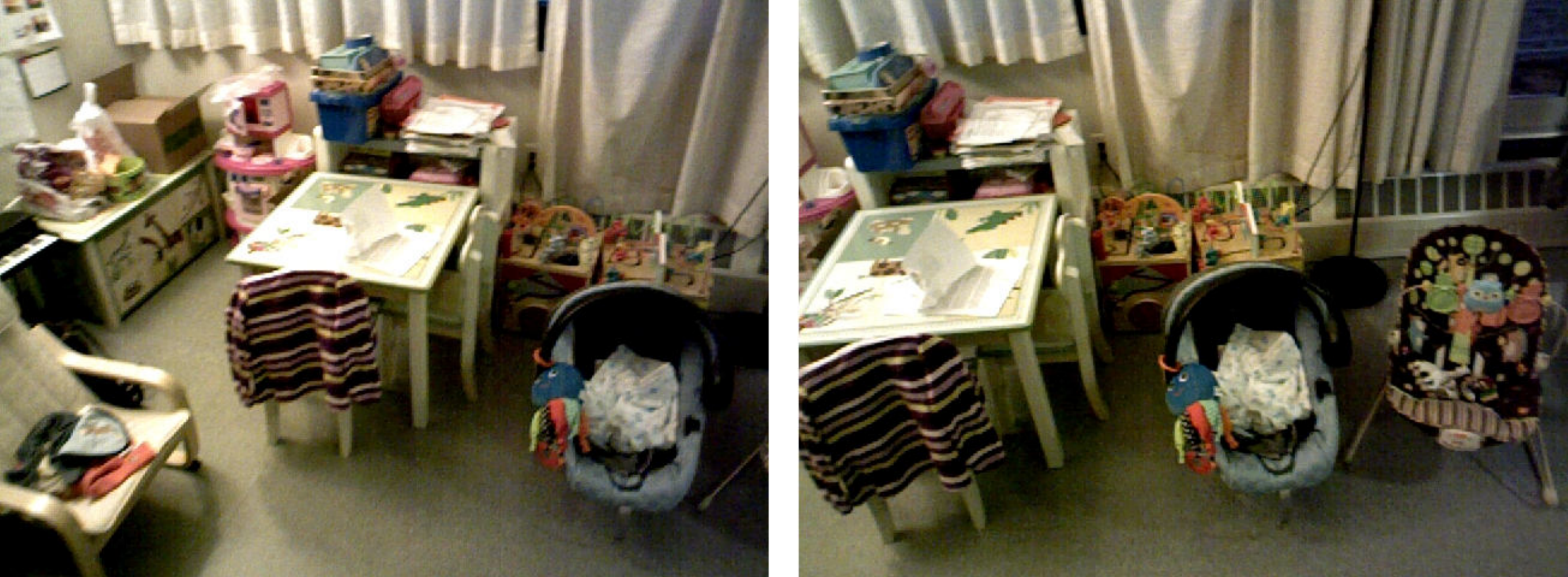}
    } 
    \subcaptionbox*{(a) OANet++}[.245\linewidth]{
        \includegraphics[width=\linewidth]{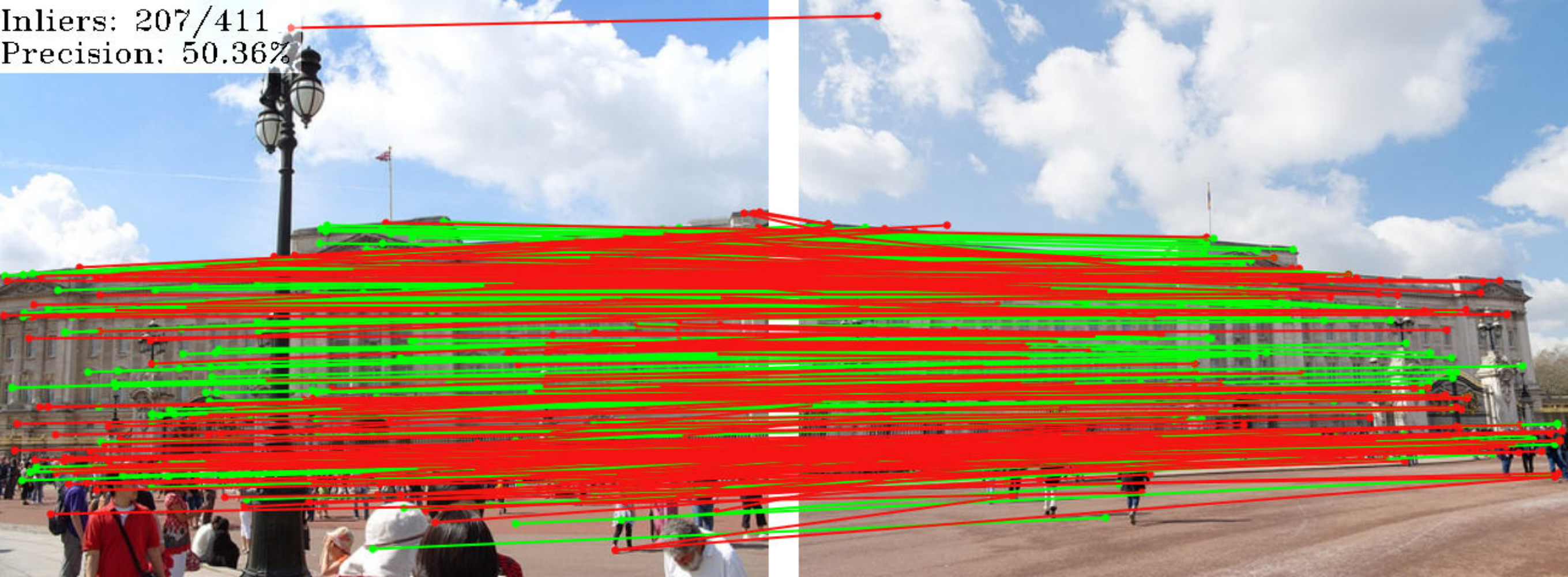}
        \includegraphics[width=\linewidth]{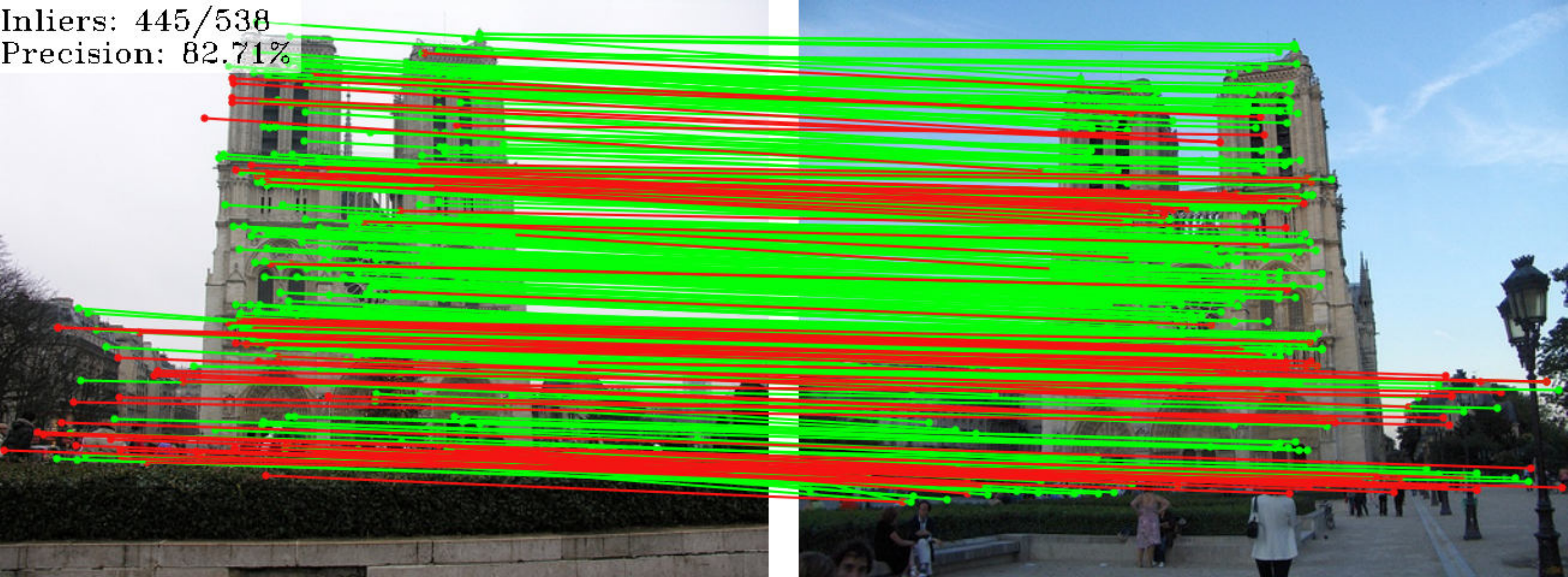}
        \includegraphics[width=\linewidth]{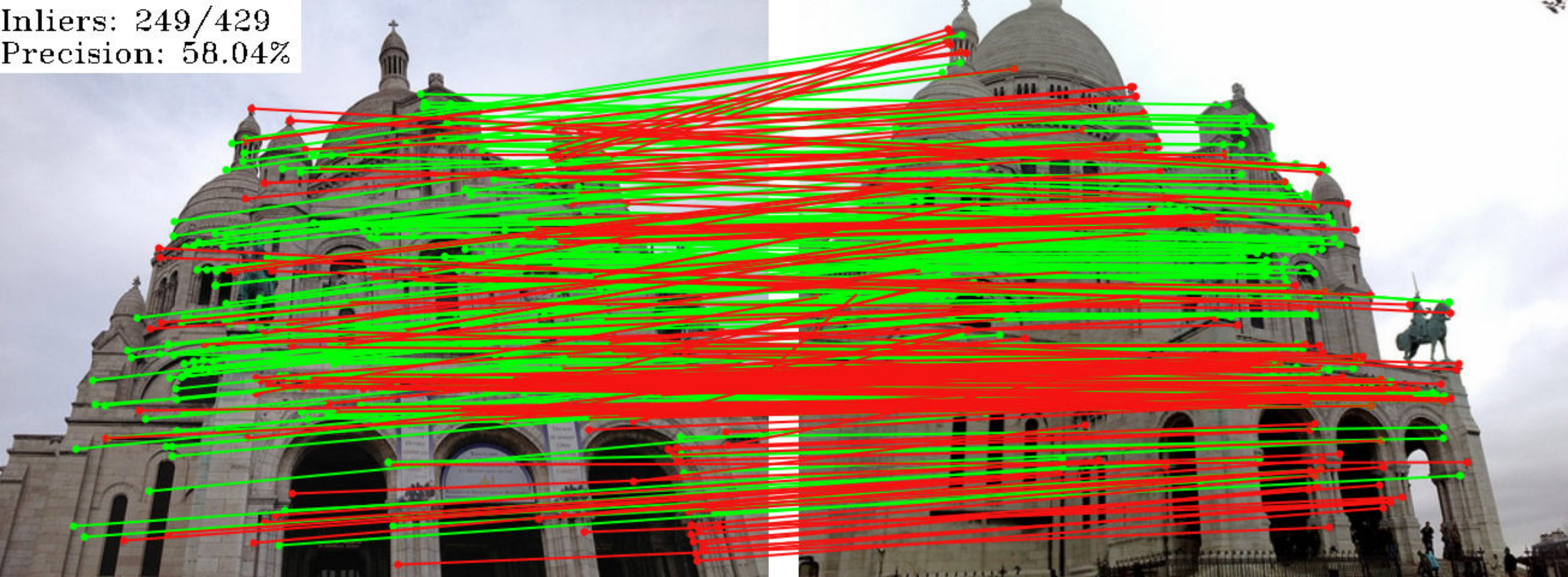}
        \includegraphics[width=\linewidth]{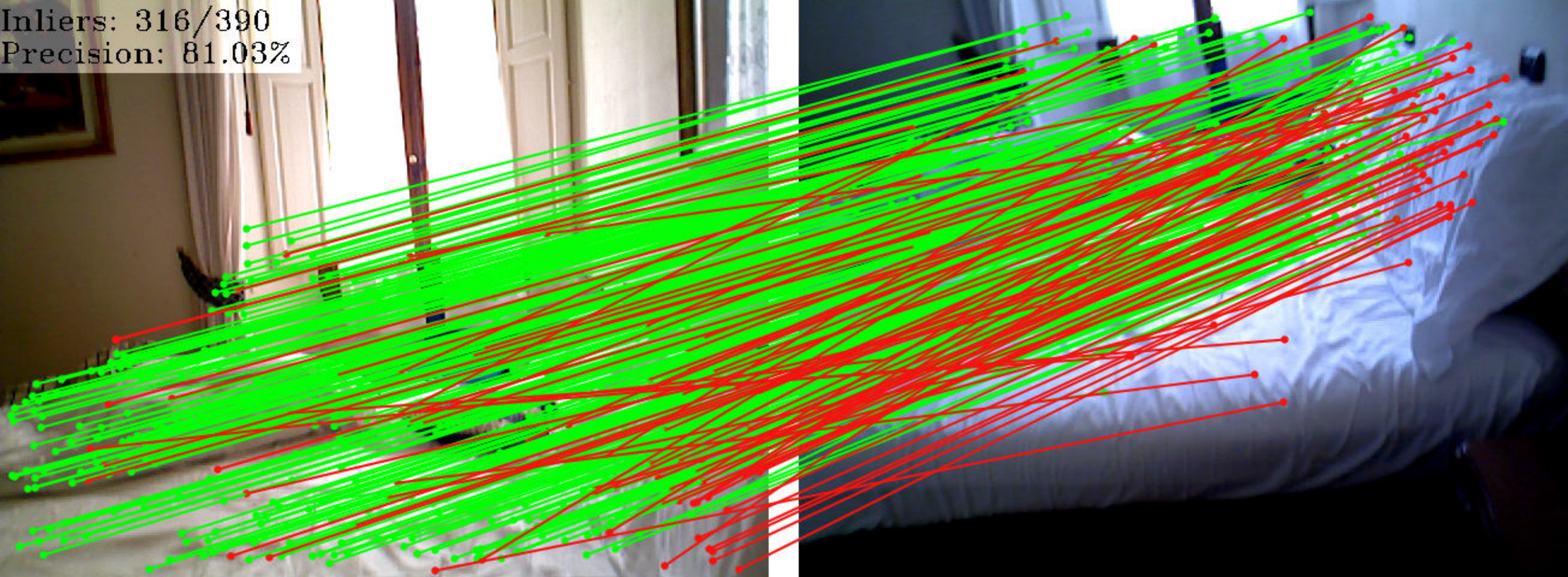}
        \includegraphics[width=\linewidth]{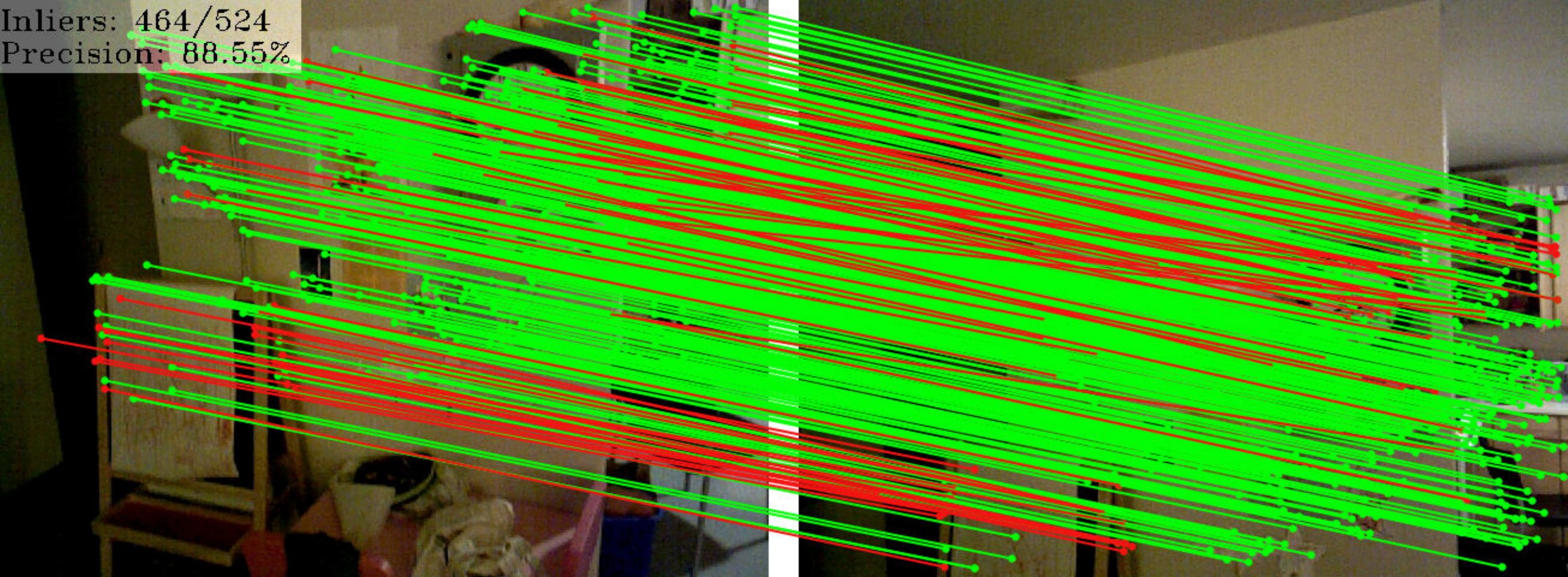}
        \includegraphics[width=\linewidth]{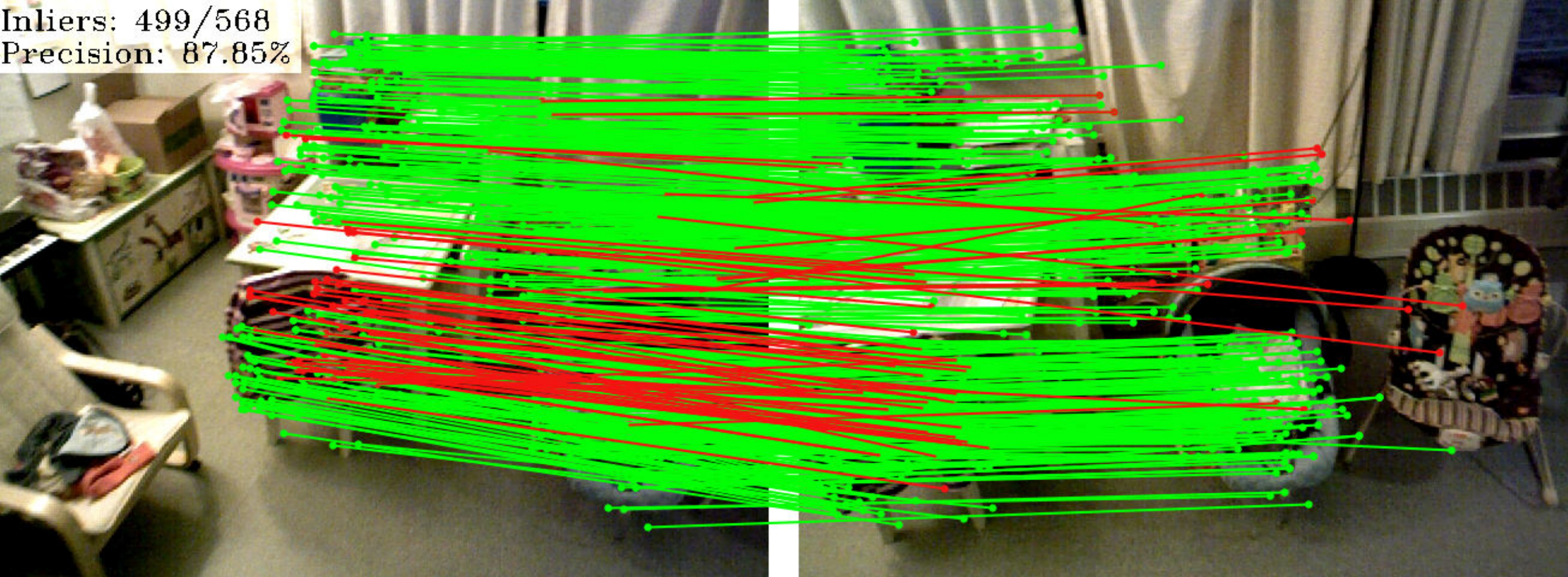}
    } 
    \subcaptionbox*{(b) CLNet}[.245\linewidth]{
        \includegraphics[width=\linewidth]{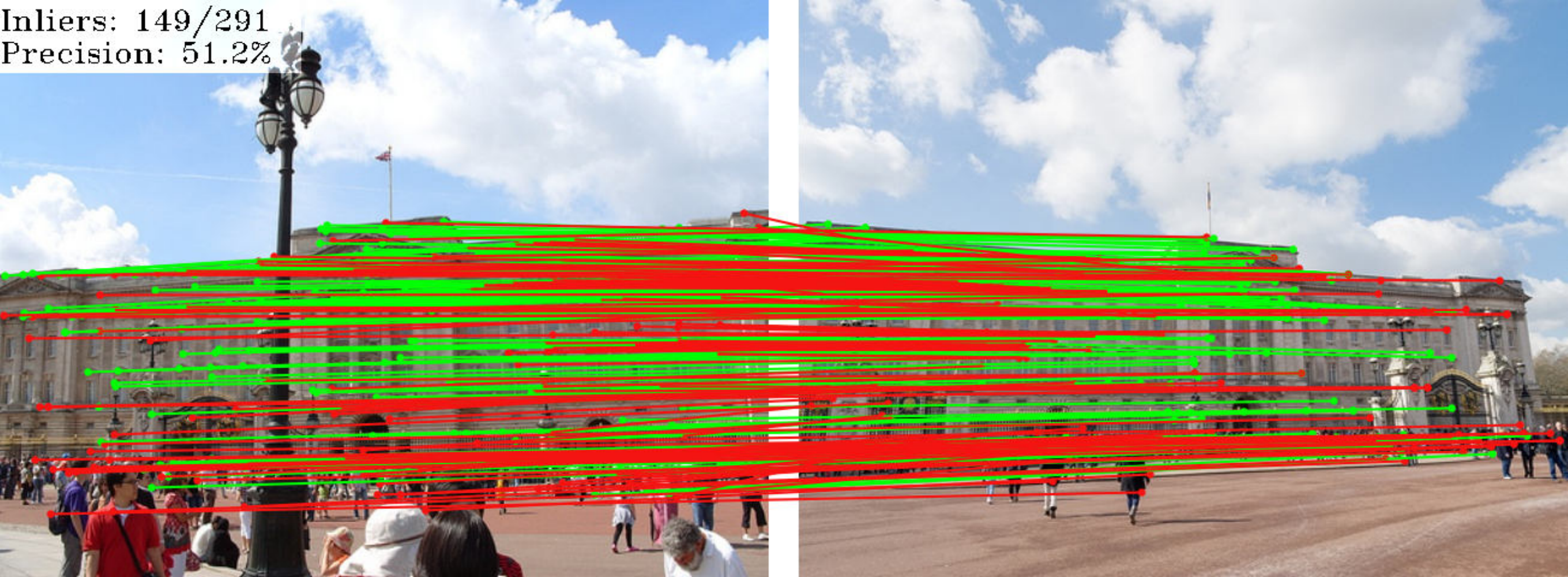}
        \includegraphics[width=\linewidth]{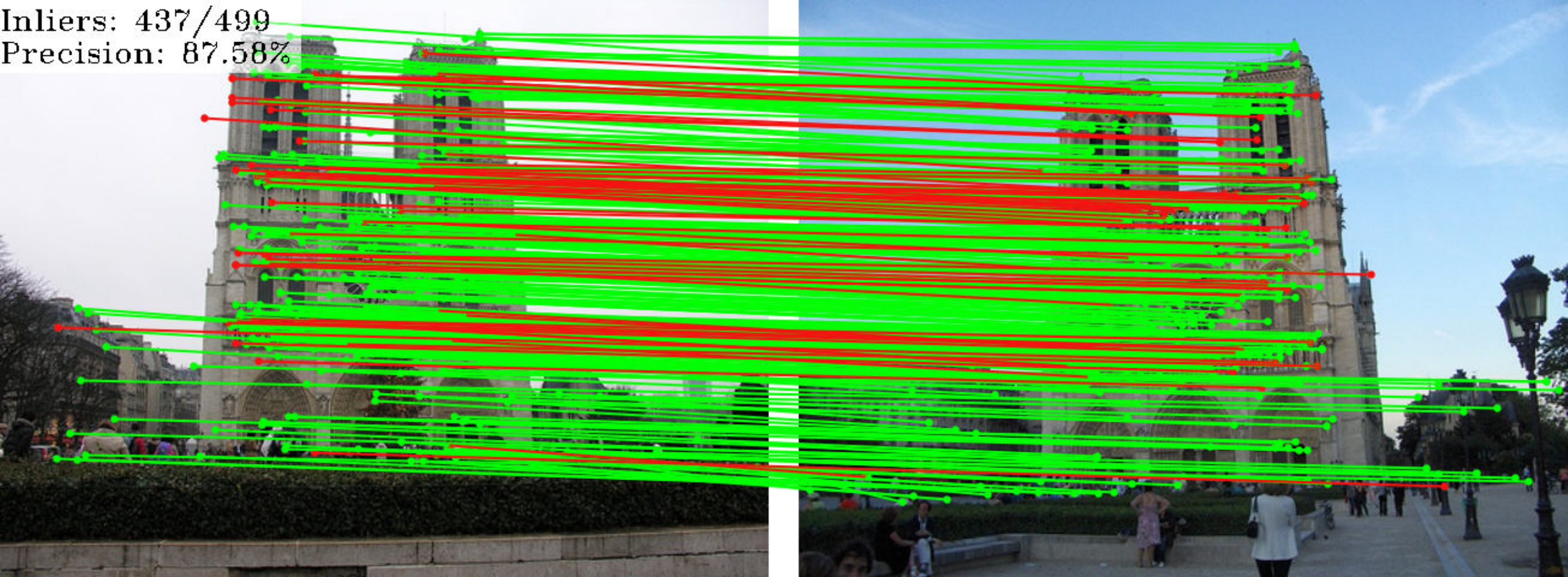}
        \includegraphics[width=\linewidth]{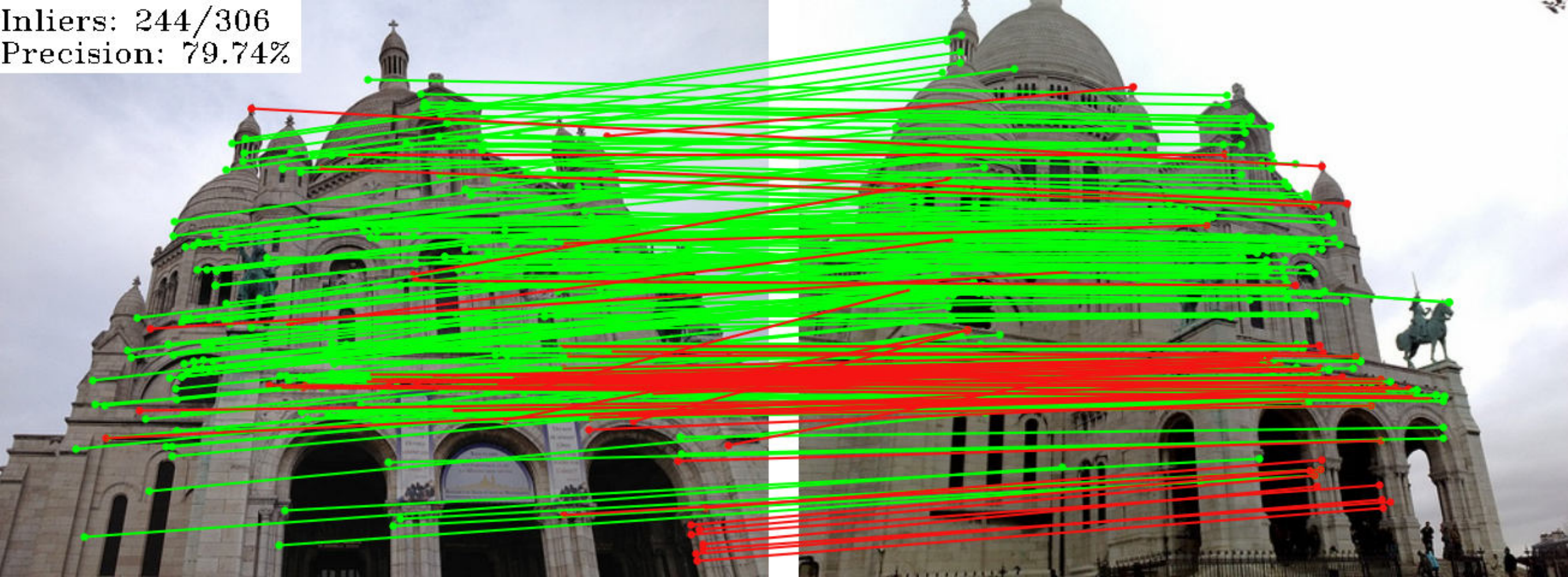}
        \includegraphics[width=\linewidth]{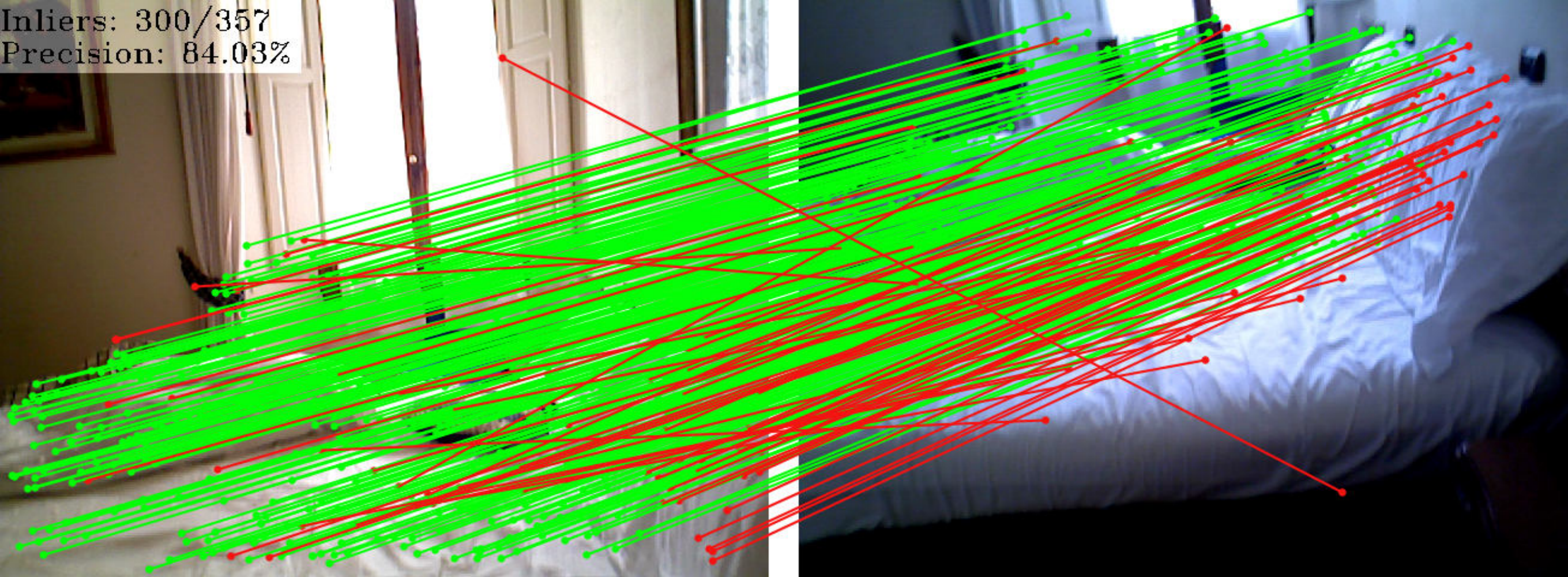}
        \includegraphics[width=\linewidth]{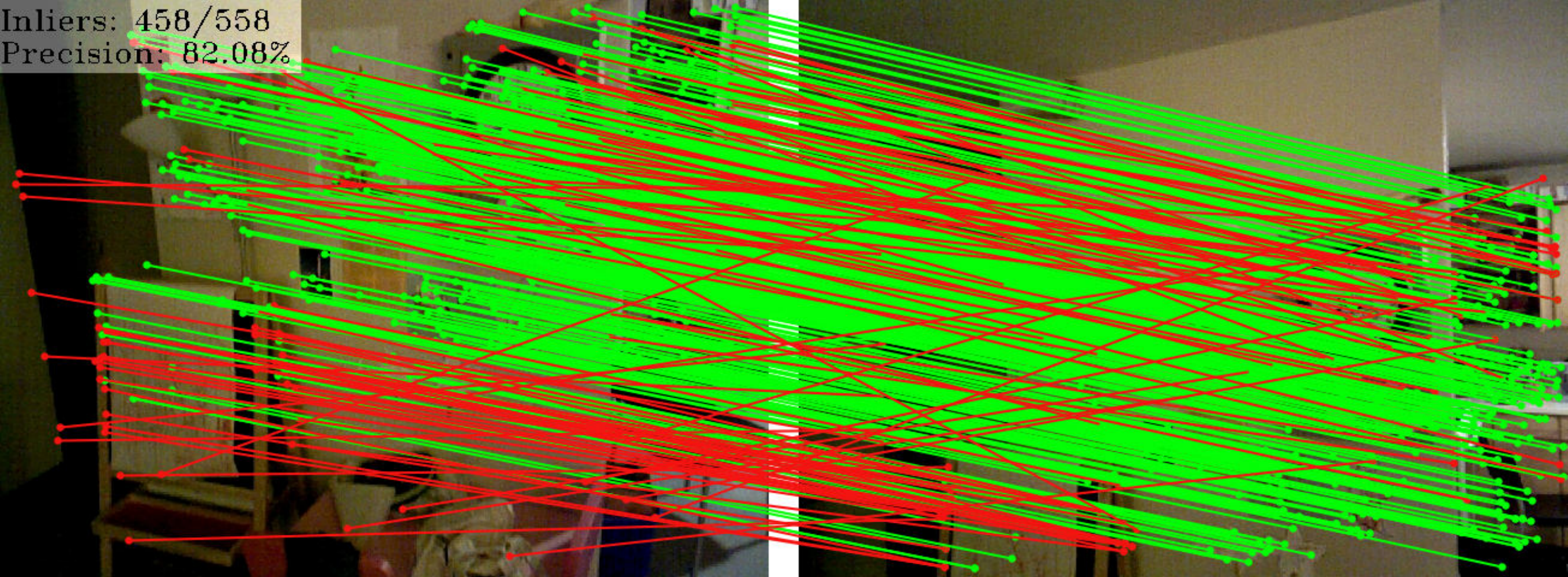}
        \includegraphics[width=\linewidth]{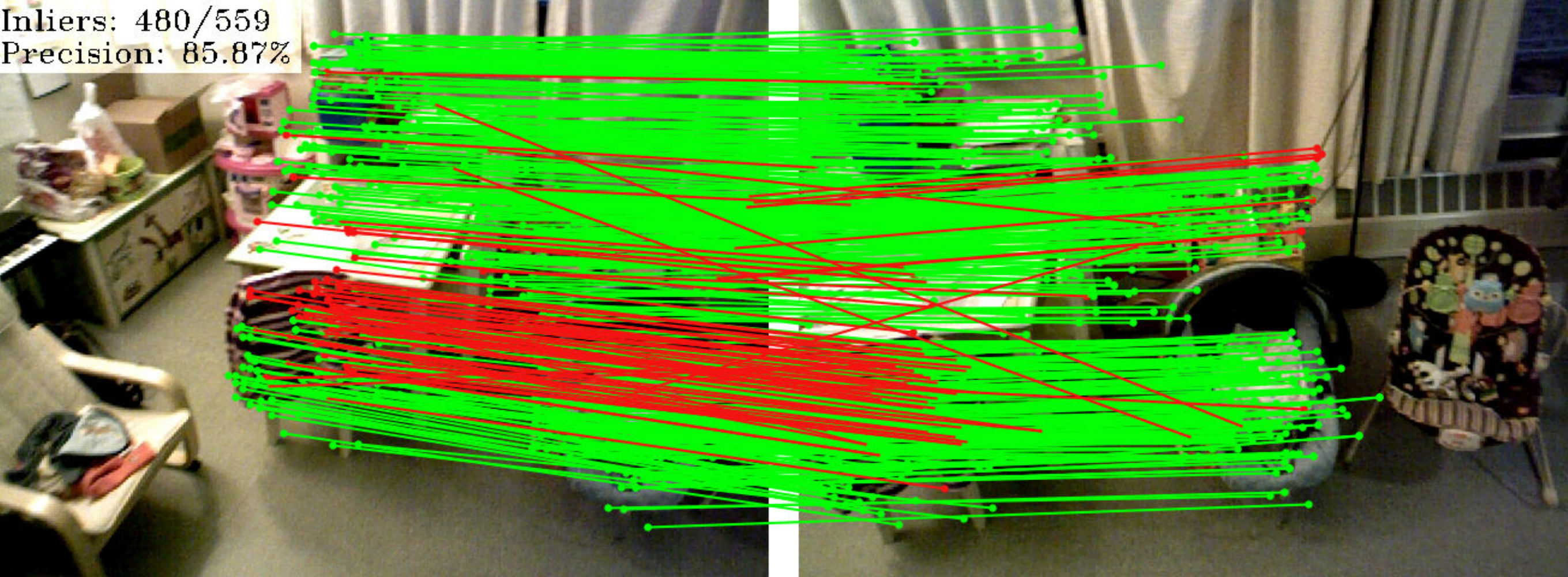}
    } 
    \subcaptionbox*{(c) Ours}[.245\linewidth]{
        \includegraphics[width=\linewidth]{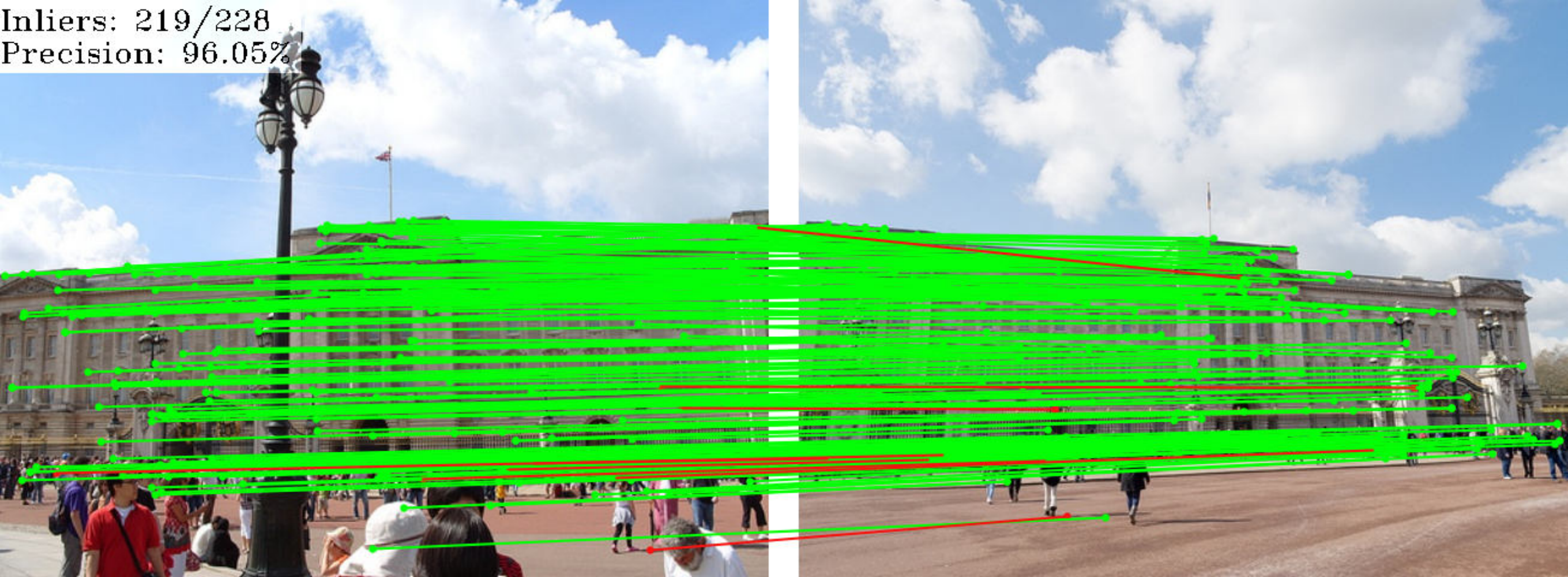}
        \includegraphics[width=\linewidth]{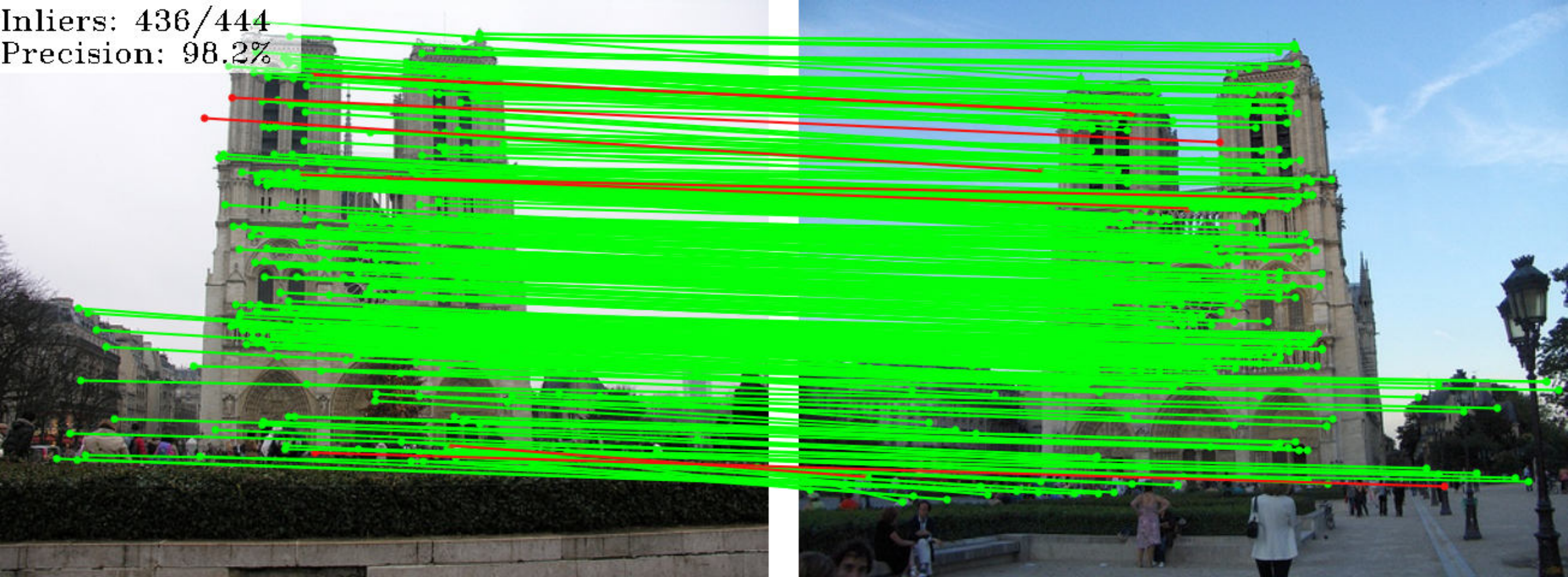}
        \includegraphics[width=\linewidth]{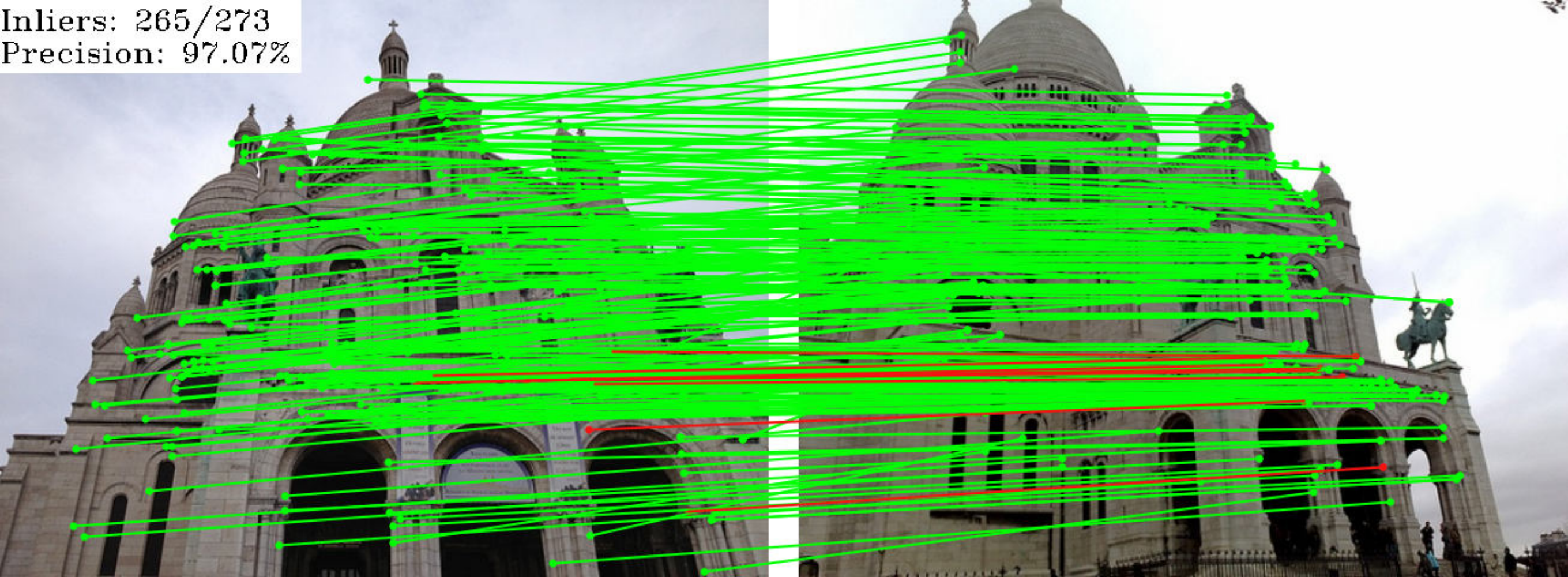}
        \includegraphics[width=\linewidth]{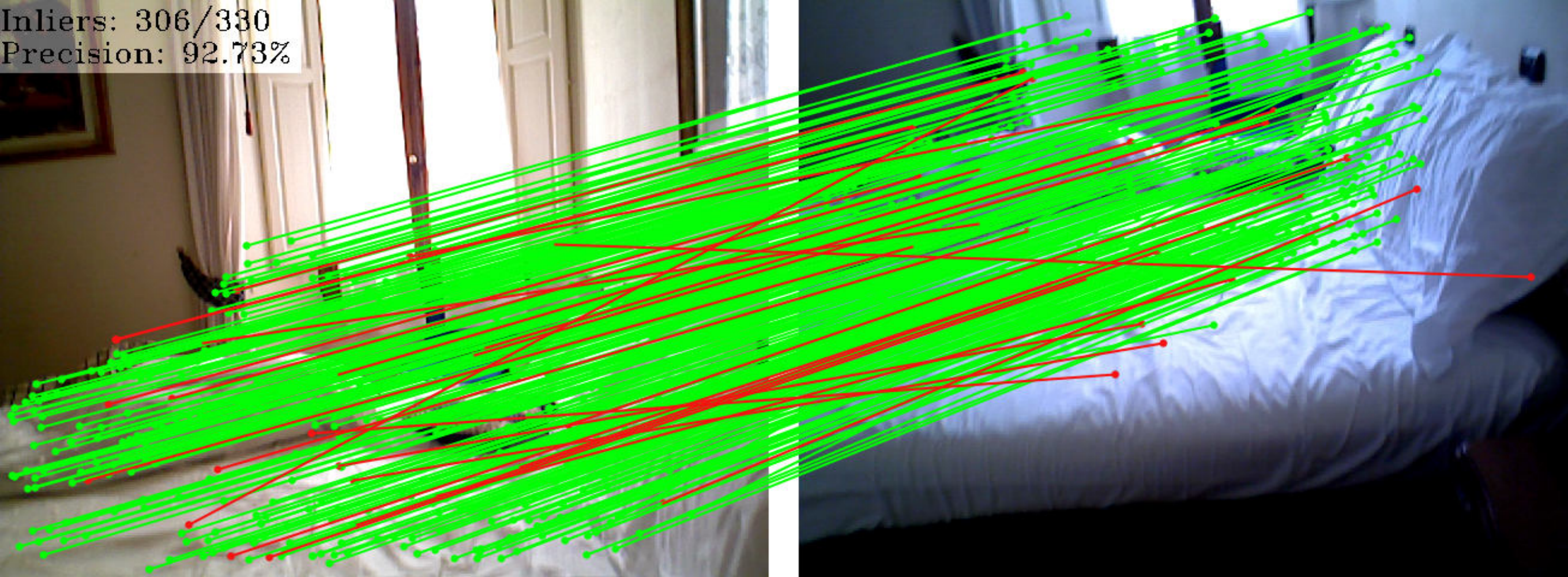}
        \includegraphics[width=\linewidth]{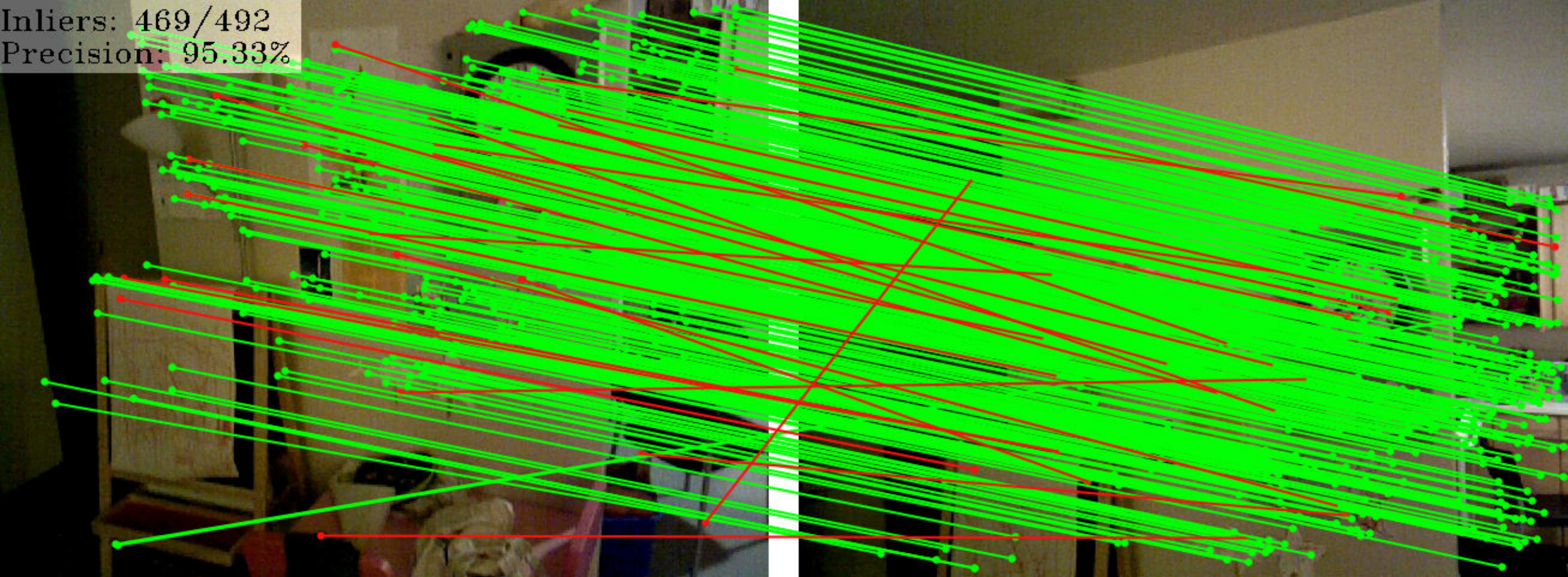}
        \includegraphics[width=\linewidth]{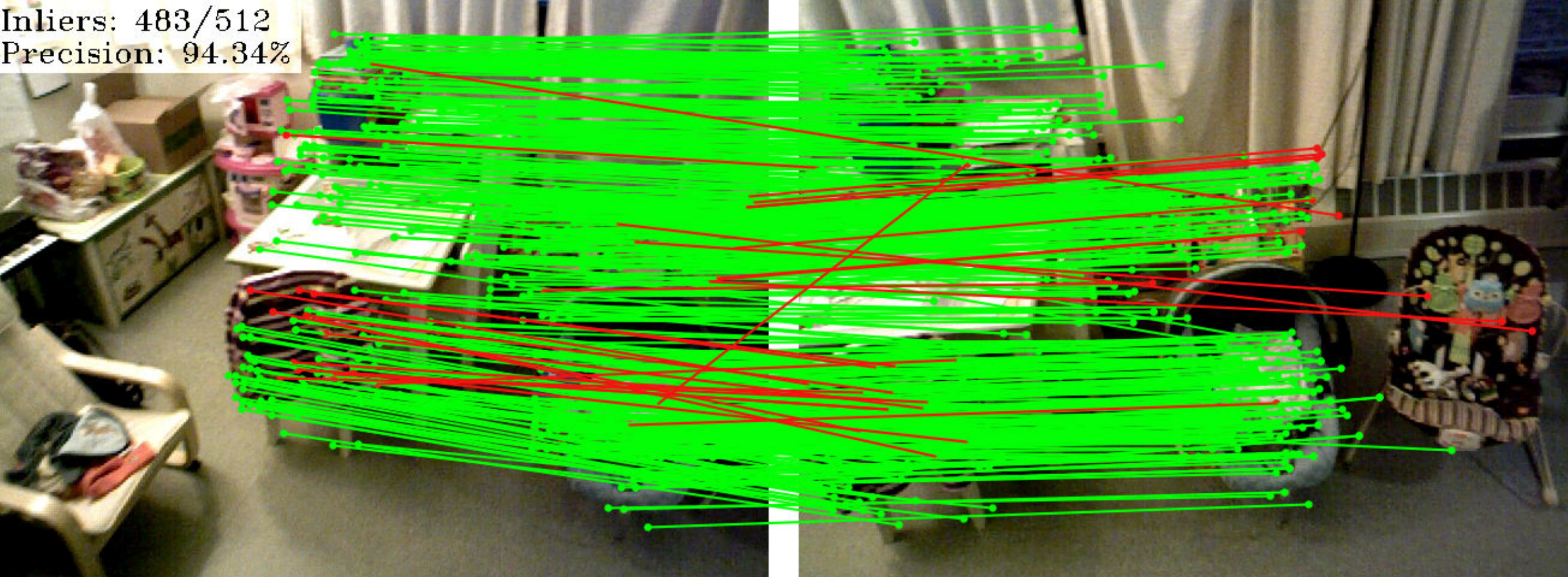}
    } 
    \caption{Partial typical visualization results on two challenging datasets, i.e., YFCC100M, SUN3D. 
    From left to right: the results of OANet++, CLNet, and our VSFormer. 
    From top to bottom: the top three results come from unknown outdoor scenes and the rest come from unknown indoor scenes. 
    The correspondence is drawn in green if it represents the true-positive and red for the false-positive. Best viewed in color.}\label{figure4}
\end{figure*}

\subsubsection{Local Context Capturer.}
Following~\cite{Wang2019, Zhao2021}, our ContextFormer first builds a KNN-based graph according to the Euclidean distances between each jointly visual-spatial correspondences: 
\begin{gather} \label{eq8}
    \mathcal{G}_i = (\mathcal{V}_i, \mathcal{E}_i), i \in \left[ 1,N \right], 
\end{gather}
where $\mathcal{V}_i = \left\{ f_{i1},...,f_{ik} \right\}$ represents the $k$-nearest neighbors of $f_i$ in the feature space; 
$\mathcal{E}_i = \left\{ \bm{e_{i1}},...,\bm{e_{ik}} \right\}$ represents the set of directed edges connecting $f_i$ and its neighbors. 
The construction of edges can be formulated as follows: 
\begin{gather} \label{eq9}
    \bm{e_{ij}} = \left[ f_i, f_i - f_{ij} \right],  
\end{gather}
where $f_i$ and $f_{ij}$ represent the $i$-th jointly visual-spatial correspondence and its $j$-th neighbor, respectively; 
$\left[ \cdot,\cdot \right]$ denotes the concatenate operation along the channel dimension. 

Then, the global graph $\mathcal{G}_{in}=\left\{ \mathcal{G}_1,...,\mathcal{G}_i,...,\mathcal{G}_N \right\}$ has rich contextual information, but capturing them only by neighborhood aggregation is not robust. 
To this end, a novel graph attention block adopts the squeeze-and-excitation mechanism to efficiently capture the potential spatial-, channel-, and neighborhood-wise relations inside the global graph. 
To be specific, as shown in Fig.~\ref{figure3}, the global graph $\mathcal{G}_{in}$ is embedded by a PointCN block~\cite{Yi2018}; 
then, the max-pooling and average-pooling operations along channel dimension to squeeze the global graph, and the element-wise summation operation is applied to produce an initial attention map $\mathbf{A_1} \in \mathbb{R}^{N \times K}$; 
subsequently, the initial attention map is excited with an MLP to capture neighborhood-wise relations $\mathbf{A_2} \in \mathbb{R}^{N \times K}$ of the global graph. 
finally, these relationships are embedded into the global graph via Hadamard product, adding a residual to obtain $\mathcal{G}_{out}$. 
The above operations can be described as: 
\begin{gather} \label{eq10}
    \mathcal{G}_{in}^{'} = \mathrm{PointCN}(\mathcal{G}_{in}), \\
    \mathbf{A_2} = \mathrm{MLP}(\mathrm{AvgPool}(\mathcal{G}_{in}^{'}) + \mathrm{MaxPool}(\mathcal{G}_{in}^{'})), \\
    \mathcal{G}_{out} = (\mathcal{G}_{in}^{'} \odot \mathbf{A_2}) + \mathcal{G}_{in}. 
\end{gather}
Similar to the neighborhood-wise attention (as described above), the operations of channel- and spatial-wise attention are omitted for simplicity. 
Despite its simplicity, the graph attention block effectively improves the representation ability of the global graph. 

Next, following~\cite{Zhao2021}, we perform neighborhood aggregation on the enhanced global graph $\mathcal{G}_{out}$ to obtain the correspondence feature $\mathbf{F_{local}} \in \mathbb{R}^{N \times C}$ embedded with both global and local contexts. 

\subsubsection{Global Context Capturer.}
It mainly involves a multi-headed self-attention (MHSA) layer employed to capture and fuse global context into each correspondence. 
In particular, this paper introduces length similarity~\cite{Bai2021} into the MHSA layer, which produces a spatially aware attention matrix by combining length consistency and feature consistency. 
To be specific, given two correspondences $\mathbf{c_i=(p_{i}^{A}, p_{i}^{B})}$ and $\mathbf{c_j=(p_{j}^{A}, p_{j}^{B})}$, the length similarity between them is computed as: 
\begin{gather} \label{eq12}
    m_{i,j} =  \big| \Vert \mathbf{p_{i}^{A} - p_{j}^{A}} \Vert - \Vert \mathbf{p_{i}^{B} - p_{j}^{B}} \Vert \big|. 
\end{gather}
Then, the length consistency is fused into the attention matrix $\mathbf{A_3} \in \mathbb{R}^{N \times N}$ by the MHSA layer, while generating a spatial aware attention matrix $\mathbf{A_3} \in \mathbb{R}^{N \times N}$ to guide message passing. 
This operation can be formulated as follows: 
\begin{gather} \label{eq13}
    \mathbf{A_4} = \mathbf{A_3} \odot \mathbf{M_{ls}}, 
\end{gather}
where $\mathbf{M_{ls}} \in \mathbb{R}^{N \times N}$ represents the length similarity matrix obtained by Eq.~\ref{eq12}. 
Besides, the other operation details are similar to those of the transformer in VSFusion, thus this paper omits the rest for simplicity. 
Finally, we adopt the same inlier predictor as~\cite{Zhao2021} to process the concatenated feature. 

\subsection{Loss Function}
Following~\cite{Hartley2003, Ranftl2018}, a hybrid loss function is employed to supervise the training process of our proposed method: 
\begin{equation} \label{eq14}
    \mathcal{L} = \mathcal{L}_c(o_j, y_j) + \alpha \mathcal{L}_e(\mathbf{\widehat{E}}, \mathbf{E}), 
\end{equation}
where $\mathcal{L}_c$ denotes the classification loss; 
$\mathcal{L}_e$ represents the essential matrix loss; 
$\alpha$ is a hyper-parameter to balance these two losses. 

Following~\cite{Zhao2021}, the classification loss $\mathcal{L}_c$ can be formulated as: 
\begin{equation} \label{eq15}
    \mathcal{L}_c(o_j, y_j) = \sum_{j=1}^{\lambda} H({\omega}_j \odot o_{j}, y_j), \\
\end{equation}
where $H(\cdot)$ denotes a binary cross entropy loss function; 
$o_j$ represents the relevant weights of the $j$-th iteration; 
$y_j$ represents the weakly supervised labels, which are chosen under the epipolar distance threshold $10^{-4}$ as positive~\cite{Hartley2003}; 
${\omega}_j$ is an adaptive temperature vector, and $\odot$ represents the Hadamard product. 

Following~\cite{Zhang2019}, the essential matrix loss $\mathcal{L}_e$ can be formulated as: 
\begin{equation} \label{eq16}
    \mathcal{L}_e = \frac{(\mathbf{p^{\prime T} \widehat{E} p})^2} {{\left\lVert\mathbf{{Ep}}\right\rVert}_{[1]}^{2} + {\left\lVert\mathbf{{Ep}}\right\rVert}_{[2]}^{2} + {\left\lVert{\mathbf{Ep^{\prime}}}\right\rVert}_{[1]}^{2} + {\left\lVert{\mathbf{Ep^{\prime}}}\right\rVert}_{[2]}^{2}},
\end{equation}
where $\mathbf{E}$ denotes the ground truth of the essential matrix; 
$\mathbf{p}$ and $\mathbf{p^{\prime}}$ represent virtual correspondence coordinates obtained by the essential matrix $\mathbf{E}$; 
${\left\lVert{\cdot}\right\rVert}_{[i]}$ denotes the $i$-th element of vector. 

\begin{table}[t]
    \tabcolsep=1.8pt 
    \renewcommand\arraystretch{1.1} 
    \centering
    \small
    \begin{tabular}{c|c||cccc}
    \hline
    \multirow{2}{*}{Method} & \multirow{2}{*}{Params.} & \multicolumn{2}{c}{YFCC100M (\%)} & \multicolumn{2}{c}{SUN3D (\%)}\\
     \cline{3-6}
     &  & - & RANSAC & - & RANSAC \\ 
    \hline \hline
    PointNet++~\shortcite{Qi2017} & 12.00M & 16.48 & 46.25 & 8.10 & 15.29\\
    PointCN~\shortcite{Yi2018} & 0.39M & 23.95 & 48.03 & 9.30 & 16.40\\ 
    DFE~\shortcite{Ranftl2018} & 0.40M & 30.27 & 51.16 & 12.06 & 16.26\\ 
    OANet++~\shortcite{Zhang2019} & 2.47M & 38.95 & 52.59 & 16.18 & 17.18\\ 
    ACNe~\shortcite{Sun2020} & 0.41M & 33.06 & 50.89 & 14.12 & 16.99\\  
    T-Net~\shortcite{Zhong2021} & 3.78M & 48.20 & 55.85 & 17.24 & 17.57\\ 
    LMCNet~\shortcite{Liu2021} & 0.93M & 47.50 & 55.03 & 16.82 & 17.38\\
    CLNet~\shortcite{Zhao2021} & 1.27M & 51.90 & 59.15 & 15.83 & 18.99\\ 
    MSA-Net~\shortcite{Zheng2022} & 1.45M & 50.65 & 56.28 & 16.86 & 17.79\\ 
    MS$^2$DG-Net~\shortcite{Dai2022} & 2.61M & 49.13 & 57.68 & 17.84 & 17.79\\
    PGFNet~\shortcite{Liu2023} & 2.99M & 53.70 & 57.83 & 19.32 & 18.00\\  
    \rowcolor{lightgray!60} Ours & 2.57M & \textbf{62.18} & \textbf{63.35} & \textbf{20.18} & \textbf{20.27}\\ 
    \hline
    \end{tabular}
    \caption{Quantitative comparison results of the camera pose estimation on unknown scenes. The mAP5\textdegree~without/with RANSAC as a post-processing step is reported.} \label{comparison1}
\end{table}

\begin{table}[t]
    \tabcolsep=2pt 
    \renewcommand\arraystretch{1.1} 
    \centering
    \small
    \begin{tabular}{c||cccc}
    \hline
    \multirow{2}{*}{Method} & \multicolumn{2}{c}{YFCC100M (\%)} & \multicolumn{2}{c}{SUN3D (\%)}\\
     \cline{2-5}
     & - & RANSAC & - & RANSAC \\ 
    \hline \hline
    PointNet++~\shortcite{Qi2017} & 10.49 & 33.78 & 10.58 & 19.17\\ 
    PointCN~\shortcite{Yi2018} & 13.81 & 34.55 & 11.55 & 20.60\\ 
    OANet++~\shortcite{Zhang2019} & 32.57 & 41.53 & 20.86 & 22.31\\ 
    ACNe~\shortcite{Sun2020} & 29.17 & 40.32 & 18.86 & 22.12\\ 
    T-Net~\shortcite{Zhong2021} & 42.99 & 45.25 & 22.38 & 22.96\\ 
    LMCNet~\shortcite{Liu2021} & 33.73 & 40.39 & 19.92 & 21.79\\ 
    CLNet~\shortcite{Zhao2021} & 38.75 & 44.88 & 19.20 & 23.83\\ 
    MSA-Net~\shortcite{Zheng2022} & 39.53 & 44.57 & 18.64 & 22.03\\ 
    MS$^2$DG-Net~\shortcite{Dai2022} & 38.36 & 45.34 & 22.20 & 23.00\\ 
    PGFNet~\shortcite{Liu2023} & 44.20 & 46.28 & 23.66 & 23.87\\ 
    \rowcolor{lightgray!60} Ours & \textbf{48.83} & \textbf{49.03} & \textbf{24.81} & \textbf{24.76}\\ 
    \hline
    \end{tabular}
    \caption{Quantitative comparison results of the camera pose estimation on known scenes. The mAP5\textdegree~without/with RANSAC as a post-processing step is reported.} \label{comparison2}
\end{table}

\subsection{Implementation Details} 
Holistically, SIFT~\cite{Lowe2004} is adopted to establish $N=2000$ initial correspondences, channel dimension $C$ is $128$, network iteration $\lambda$ is $2$, and pruning ratio $r$ is $0.5$; 
besides, considering reducing the training cost, this paper only uses VSFusion in the second iteration. 
In VCExtractor, the original images are resized to $H = 120, W = 160$, and the channel dimension $C_F$ is $64$. 
In ContextFormer, the number of $k$ neighbors is set to $9$ and $6$ for two iterations. 
We adopt Adaptive Moment Estimation (Adam) with a weight decay of $0$ as the optimizer to train our network, and the canonical learning rate (for batch size is $32$) is set to $10^{-3}$. 
Following~\cite{Zhao2021}, the weight $\alpha$ in Equation~\ref{eq14} is set as $0$ during the first $20k$ iterations and $0.5$ in the remaining $480k$ iterations. 

\section{Experiments and Analysis}
\label{sec:experiment}
\subsection{Evaluation Protocols}
We test our method on both outdoor and indoor benchmarks to evaluate the performance on relative pose estimation~\cite{Zhang2019}. 
Yahoo's YFCC100M~\cite{Thomee2016} dataset is used as our outdoor scenes, which is made up of 100 million outdoor photos from the Internet. 
The SUN3D~\cite{Xiao2013} dataset is used as our indoor scenes, which consists of large-scale indoor RGB-D videos with information about camera poses. 
Following the data division of~\cite{Zhang2019}, all methods are evaluated on both unknown scenes and known scenes. 
In this paper, the mAP of pose error at the thresholds (5\textdegree~and 20\textdegree) are reported, where the pose error is the maximum of angular errors from rotation and translation. 

\begin{table}[t]
    \tabcolsep=2pt 
    \renewcommand\arraystretch{1.1}
    \centering
    \small
    \begin{tabular}{ccccc||cc}
    \hline
    ResNet & VSFusion & ContextFormer & 2x & 3x & mAP5\textdegree & mAP20\textdegree\\
    \hline \hline
    $\checkmark$ &  &  & $\checkmark$ &  & 43.25 & 67.12\\
    $\checkmark$ & $\checkmark$ &  & $\checkmark$ &  & 49.63 & 72.87\\
     &  & $\checkmark$ & $\checkmark$ &  & 57.55 & 78.12\\
     & $\checkmark$ & $\checkmark$ & $\checkmark$ &  & \textbf{62.18} & \textbf{80.95}\\
     & $\checkmark$ & $\checkmark$ &  & $\checkmark$ & 55.70 & 76.08\\
    \hline
    \end{tabular}
    \caption{Ablation study of network architecture. 
    2x and 3x represent the number of network iterations. }\label{ablation1}
\end{table}

\subsection{Comparison Results} 
As shown in Table~\ref{comparison1} and Table~\ref{comparison2}, our VSFormer outperforms other state-of-the-art (SOTA) methods on outdoor and indoor scenes. 
To be specific, on outdoor scenes, our method achieves a performance improvement of 15.8\% over the recent MLP-based SOTA method (PGFNet) on unknown scenes at mAP5\textdegree~without RANSAC. 
Similarly, compared to the recent Graph-based SOTA method (CLNet), our method attains a performance improvement of 19.8\% at mAP5\textdegree~without RANSAC. 
On indoor scenes, our method achieves a performance improvement of 24.7\% and 27.5\% over two baselines (OANet++ and CLNet) on unknown scenes at mAP5\textdegree~without RANSAC, respectively. 
Meanwhile, our method also achieves the best performance among all methods with RANSAC. 
The results indicate that our proposed visual-spatial fusion and transformer-based structure can further improve the network performance. 
Additionally, as shown in Fig.~\ref{figure4}, partial typical visualization results of OANet++~\cite{Zhang2019}, CLNet~\cite{Zhao2021}, and our network are shown from left to right. 
It can be seen that our method achieves the best performance under various challenging scenes. 

\subsection{Ablation Studies}
To deeply analyze the proposed method, we perform detailed ablation studies on YFCC100M to demonstrate the effectiveness of each component in VSFormer. 

\subsubsection{Network Architecture.}

As shown in Table~\ref{ablation1}, we intend to gradually add these components to the baseline. 
The baseline (Row-1) we used is PointCN~\cite{Yi2018} with the pruning strategy. 
It can be seen that all our component combinations outperform the baseline on outdoor scenes. 
To be specific, in the second row, the VSFusion is first introduced, which achieves a performance improvement of 14.8\% over the baseline at mAP5\textdegree~without RANSAC. 
It indicates the importance of exploiting visual cues of a scene/image pair to guide correspondence pruning. 
Meanwhile, as illustrated in the third row, replacing ResNet encoders~\cite{Yi2018} with our ContextFormer, 
which obtained a performance improvement of 33.1\% over the baseline at mAP5\textdegree~without RANSAC. 
Moreover, when combining the proposed modules and using two iterations, the mAP5\textdegree~is significantly better than those of the baselines. 
As shown in the last row, this paper also explores the effect of increasing the number of network iterations. 
The experiment demonstrates that this leads to a performance penalty of 10.4\% over the proposed method. 
This is mainly because the redundant iterations discard some inliers, which are important for the geometric model estimation. 
That is, the task of camera pose estimation requires sufficient and accurate inliers. 

\begin{table}[t]
    \tabcolsep=2pt 
    \renewcommand\arraystretch{1.1}
    \centering
    \small
    \begin{tabular}{ccccc||cc}
    \hline
    ContextFormer & Cross & TR & Proj & Sum & mAP5\textdegree & mAP20\textdegree\\
    \hline \hline
    $\checkmark$ &  &  &  &  & 57.55/61.82 & 78.12/81.20\\
    $\checkmark$ & $\checkmark$ &  &  &  & 60.33/61.88 & 80.04/81.19\\
    $\checkmark$ &  & $\checkmark$ &  &  & 60.18/61.68 & 79.89/81.24\\
    $\checkmark$ &  & $\checkmark$ & $\checkmark$ &  & 61.23/62.43 & 80.60/81.20\\
    $\checkmark$ &  & $\checkmark$ & $\checkmark$ & $\checkmark$ & \textbf{62.18/63.35} & \textbf{80.95/81.84}\\
    \hline
    \end{tabular}
    \caption{Ablation study for the VSFusion. The results of mAP (\%) with/without RANSAC on unknown scenes are reported. Cross, TR, Proj, and Sum respectively represent the cross-attention layer, transformer layer, projection layer, and final element-wise summation.}\label{ablation2}
\end{table}

\begin{figure}[t]
    \centering
    \includegraphics[width=1\linewidth]{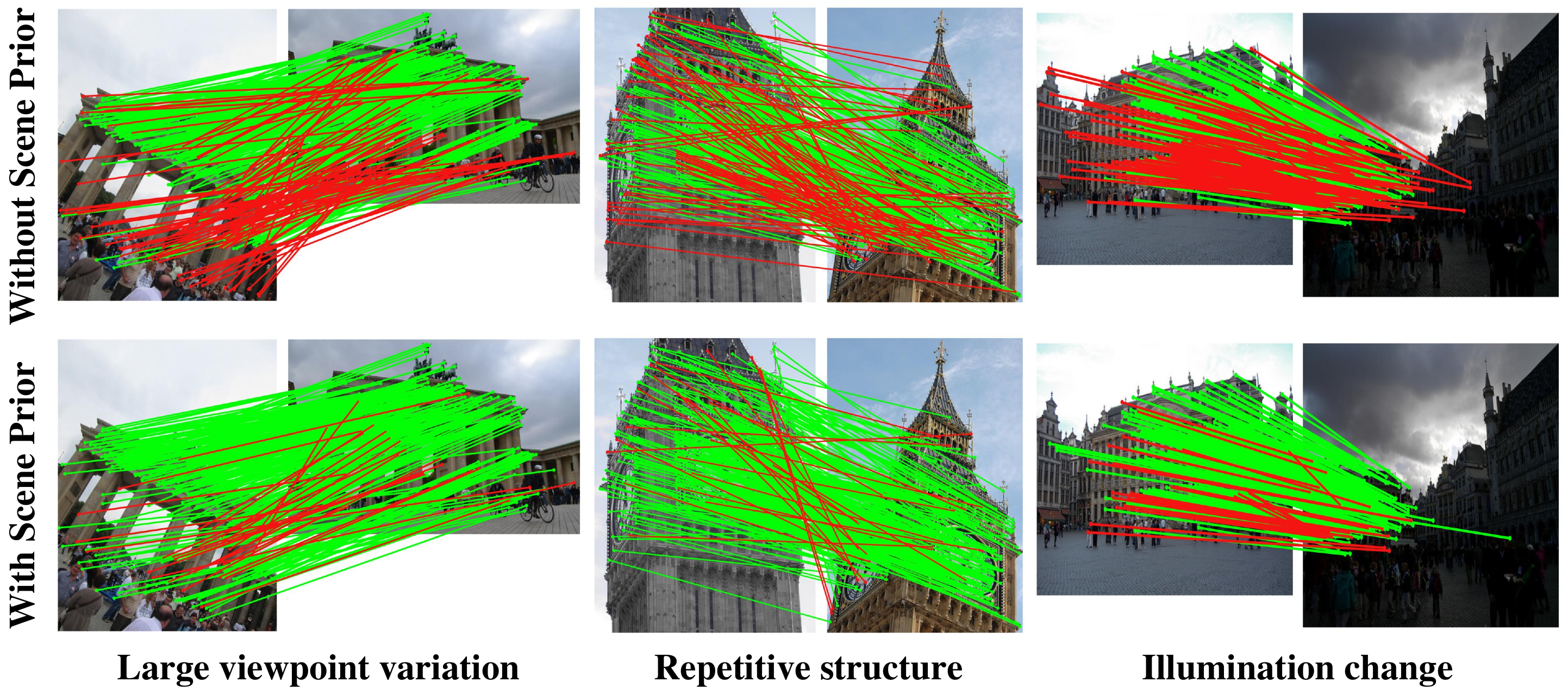}
    \caption{Qualitative comparison with/without scene prior. }\label{figure5}
\end{figure}

\begin{table}[t]
    \tabcolsep=2pt 
    \renewcommand\arraystretch{1.}
    \centering
    \small
    \begin{tabular}{ccccc}
    \toprule
    \multirow{2}{*}{Method} & \multicolumn{2}{c}{Known Scene (\%)} & \multicolumn{2}{c}{Unknown Scene (\%)}\\
    \cmidrule(r){2-3} \cmidrule(r){4-5}
     & mAP5\textdegree & mAP20\textdegree & mAP5\textdegree & mAP20\textdegree\\ 
    \midrule
    PointCN & 13.81 & 35.20 & 23.95 & 52.44 \\ 
     \rowcolor{lightgray!60} PointCN* & \textbf{24.87}$_{+11.06}$ & \textbf{47.96}$_{+12.76}$ & \textbf{28.18}$_{+4.23}$ & \textbf{56.57}$_{+4.13}$\\ 
     OANet++ & 32.57 & 56.89 & 38.95 & 66.85\\
     \rowcolor{lightgray!60} OANet++* & \textbf{37.90}$_{+5.33}$ & \textbf{59.97}$_{+3.08}$ & \textbf{46.10}$_{+7.15}$ & \textbf{70.68}$_{+3.83}$\\ 
     CLNet & 38.27 & 62.48 & 51.80 & 75.76\\ 
     \rowcolor{lightgray!60} CLNet* & \textbf{40.58}$_{+2.31}$ & \textbf{63.06}$_{+0.58}$ & \textbf{55.20}$_{+3.40}$ & \textbf{76.83}$_{+1.07}$\\ 
    \bottomrule
    \end{tabular}
    \caption{Quantitative comparison on outdoor scenes without RANSAC. The performance of the baseline can be comprehensively improved after using VSFusion. } \label{costfree}
\end{table}

\subsubsection{VSFusion Module.}
As shown in Table~\ref{ablation2}, we explore the detailed design of VSFusion. 
Comparing the results of Row-2 and Row-5, we can see that our VSFusion can achieve better performance than using a simple cross-attention layer to fuse visual and spatial cues. 
Experiments in the fourth and fifth rows verify the necessity of spatial projection and re-fusion. 
As illustrated in Fig.~\ref{figure5}, visualization results in some challenging scenes also highlight the importance of scene visual cues. 
In addition, as shown in Table~\ref{costfree}, our proposed VSFusion can be used as a plug-and-play module to improve the performance of some baselines. 

\begin{table}[t]
    \tabcolsep=2pt 
    \renewcommand\arraystretch{1.1}
    \centering
    \small
    \begin{tabular}{cccccc||cc}
    \hline
    Encoder & Graph & NA & GAB & TR & LS & mAP5\textdegree & mAP20\textdegree\\
    \hline \hline
    $\checkmark$ &  &  &  &  &  & 49.63/57.90 & 72.87/77.93\\
    $\checkmark$ & $\checkmark$ &  &  &  &  & 51.53/57.78 & 74.60/78.43\\
    $\checkmark$ & $\checkmark$ & $\checkmark$ &  &  &  & 53.73/59.88 & 75.48/79.67\\
    $\checkmark$ & $\checkmark$ & $\checkmark$ & $\checkmark$ &  &  & 58.93/61.95 & 79.69/81.50\\
    $\checkmark$ & $\checkmark$ & $\checkmark$ & $\checkmark$ & $\checkmark$ &  & 59.58/62.65 & 79.68/81.48\\
    $\checkmark$ & $\checkmark$ & $\checkmark$ & $\checkmark$ & $\checkmark$ & $\checkmark$ & \textbf{62.18/63.35} & \textbf{80.95/81.84}\\
    \hline
    \end{tabular}
    \caption{Ablation study for the ContextFormer.  
    Encoder, Graph, NA, GAB, TR, and LS represent the ReNet encoder, KNN-based graph, neighborhood aggregation, graph attention block, transformer, and introduced length similarity matrix, respectively. }\label{ablation3}
\end{table}

\subsubsection{ContextFormer.}
In this paper, we also conduct some ablation studies to verify the effectiveness of each component in the proposed structure. 
As shown in Table~\ref{ablation3}, each component of the proposed ContextFormer can further improve the network performance. 
Among them, the proposed graph attention block is the core component of the structure. 
Experiments also show that our graph attention block achieves a performance improvement of 9.68\%. 
That is, the KNN-based graph has rich context information, and our method can effectively capture these potential relationships. 

\section{Conclusion}
\label{sec:conclusion}
In this paper, with another perspective, we exploit visual cues of a scene/image pair to guide correspondence pruning. 
To this end, we design a joint visual-spatial fusion module to fuse visual and spatial cues. 
Additionally, to mine consistency within correspondences, we propose a context transformer to explicitly capture both local and global contexts. 
Meanwhile, a graph attention block is designed to mine contextual information inside the KNN-based graph. 
Both comparative and ablation experiments demonstrate the effectiveness of our proposed method, which can achieve better performance with fewer parameters. 

\section{Acknowledgements}
This work was supported in part by the National Natural Science Foundation of China under Grant U2033210 and Grant 62072223 and in part by the Zhejiang Provincial Natural Science Foundation under Grant LDT23F02024F02. 

\bibliography{aaai24}

\end{document}